\RequirePackage{fix-cm}
\documentclass{svjour3}                     
\smartqed  
\usepackage{graphicx}

\usepackage{mathptmx} 
\usepackage{amsmath} 
\usepackage{times} 
\usepackage{amsmath} 
\usepackage{amssymb}  

\graphicspath{{images_new/}}

\begin{document}
	
	\title{A Deep Neural Model Of Emotion Appraisal 
	}
	

	\author{Pablo Barros         \and
		Emilia Barakova \and 
		Stefan Wermter 
	}
	
	
	\institute{Pablo Barros \at
		Knowledge Technology, Department of Informatics, \\ University of Hamburg, Hamburg, Germany \\
		\email{barros@informatik.uni-hamburg.de}           
		\and
		Emilia Barakova \at
		Dept. of Industrial Design\\
		Eindhoven University of Technology, The Netherlands\\
		\email{e.i.barakova@tue.nl}
		\and
		Stefan Wermter \at
		Knowledge Technology, Department of Informatics, \\ University of Hamburg, Hamburg, Germany \\
		\email{wermter@informatik.uni-hamburg.de}           
	}

	%
	%
	
	\date{Received: date / Accepted: date}

	\maketitle
	
	\begin{abstract}
		
		Emotional concepts play a huge role in our daily life since they take part into many cognitive processes: from the perception of the environment around us to different learning processes and natural communication.  
		Social robots need to communicate with humans, which increased also the popularity of affective embodied models that adopt different emotional concepts in many everyday tasks. However, there is still a  
		gap between the development of these solutions and the integration and development of a complex emotion appraisal system, which is much necessary for true social robots. In this paper, we propose a deep neural model which is designed in the light of different aspects of developmental learning of emotional concepts to provide an integrated solution for internal and external emotion appraisal. We evaluate the performance of the proposed model with different challenging corpora and compare it with state-of-the-art models for external emotion appraisal. To extend the evaluation of the proposed  model, we designed and collected a novel dataset based on a  
		Human-Robot Interaction (HRI) scenario. We deployed the model in an iCub robot and evaluated the capability of the robot to learn and describe the affective behavior of different persons based on observation. The performed experiments demonstrate that the proposed model is competitive with the state of the art in describing emotion behavior in general. In addition, it is able to generate internal emotional concepts that evolve through time: it continuously forms and updates the formed emotional concepts, which is a step towards creating an emotional appraisal model grounded in the robot experiences.
		%
		%
		
		\keywords{Embedded model of emotion appraisal \and Emotions in Robots \and Deep Learning \and Human-Robot Interaction}
	\end{abstract}
	
	\section{Introduction}
	\label{introduction}  
	
	The ability to perceive, understand and respond to social interaction in a human-like manner is one of the most desired skills for social robots. This set of skills is highly complex and dependent on several different research fields, including affective understanding \cite{Foroni2009}. A robot that can learn how to recognize, understand and, most importantly, adapt to different affective reactions of humans can increase its own social capabilities by being able to interact and communicate in a natural way \cite{Bandyopadhyay2013,ghazali2018influence}. Such robot would be able to develop a judgmental capability based on the social behavior of humans around it, which could be used to accomplish complex tasks such as to create a social discernment about the environment it is in, or to be able to solve a cognitive decision-level task \cite{Rani2004,Lourens2010}. However, to use affective information on such a high level of cognitive display, an appraisal model for emotion perception is necessary \cite{Ziemke2009}.
	
	The appraisal theory has its roots in psychology and describes how humans extract affect from experienced events that cause specific reactions in different people  \cite{Frijda1989}. The appraisal of a certain perceived or remembered event causes a response which is modulated by an affective interpretation of the situation at hand. This is a very important behavior for social robots, as the modulation of actions by perceived or experienced events through emotional concepts helps on the integration and acceptance of these robots within the human society, and also contributes to understandable interactions with humans\cite{Wehrle1998}.
	
	
	
	An important aspect of emotion appraisal of a robot is its capability to perceive and categorize affective behavior of humans. However, to treat emotion expressions as the final goal of a social robot, although necessary, reduces its usability in more complex scenarios \cite{Canamero2005}. To create a general affective model to be used as a modulator for different cognitive tasks, such as intrinsic motivation, creativity, grounded learning and human-level communication or emotion perception cannot be the main, and only, focus \cite{Moerland2018}. The integration of perception with intrinsic concepts of emotional behavior, such as a dynamic and evolving mood and affective memory, is mandatory to model the necessary complexity and adaptability of an effective social robot.
	
	
	This paper proposes an Emotional Deep Neural Circuitry, which is an integrative and self-adaptive emotion appraisal model that integrates crossmodal perception, emotional concept learning, affective memory and affective modulation. Our hybrid neural model is inspired by the behavioral corelates of emotion processing in humans, from perception to learning affects and emotional modulation. To accomplish that, we based our solution on deep self-organizing layers, which are integrated in different topological and conceptual levels. Our solution is separated into a perception module, representing external emotion appraisal, and an intrinsic model which formulates the internal appraisal.

	The proposed framework is suitable to be used as a robust perception model and as an adaptive long-term behavioral description system. By using different, but integrated, learning paradigms, our model has self-adaptive qualities which are necessity for social robots. Also, by using affective memory layers, the model is able to learn and describe individualized affective behavior and take this into consideration to describe a general affective scenario. This way, our framework can be used to model complex long-term behavior of different humans interacting in a real-world environment, which is a crucial part of different cognitive tasks.
	
	We evaluate the proposed model in two different sets of experiments: one to evaluate the external and internal appraisal modules and one to evaluate the use of the model in a robotic experiment for describing social behavior. To evaluate the developed appraisal modules, we use a series of different and well-known databases to measure the performance of the network when processing individual modalities. To evaluate the formation of the intrinsic mood, we could not find any existing dataset with continuous interactions and methodology for evaluation, and thus, we designed and collected our own dataset based on long-term emotion behavior in a Human-Robot Interaction (HRI) scenario and further propose an evaluation methodology. We evaluate how an iCub robot is able to describe the behavior of a person by learning through the observation of different persons interacting with each other. Finally, we present an overview of the different abstraction levels of the proposed model, and how they learn and are affected by the proposed modulatory characteristics.

	\section{Related Work}

	In the past years, much research was done in the field of emotion appraisal, but some core problems are still present: most of the published works are restricted to emotion recognition, mainly assigning emotions to a set of categorical emotions \cite{Zeng2009,Corneanu2016}. These models do not look at causal origins and the dynamics of these emotions. Although a series of works take into consideration the modeling of emotional perception, most of them do it based on the theoretical behavior of simulated agents, and not in real-world scenarios with humans \cite{Miwa2003,Xiao2016}. The unification of these two solutions would benefit a complete emotion appraisal model, which would be able to describe and understand what was perceived and how this affects an internal emotion representation. 
	
	From the perception side, several studies are based on the work of \cite{Ekman1971}, which shows that certain emotion expressions are universally understood, independent of cultural background, gender and age. They established what is known as the six universal emotions:  ``Disgust'', ``Fear'', ``Happiness", ``Surprise", ``Sadness" and ``Anger" which are popular in different affective computing models. Although it is highly accepted that these concepts are perceived in a consistent manner by different persons around the world, the concept of spontaneous expressions makes these definitions harder to be used to describe daily-life scenarios. A person will express differently, sometimes even combining different characteristics of the universal emotions, depending on the situation, the company and even the time of the day \cite{Hess2016}. Based on that, several researchers built their own complex emotional states, such as confusion, surprise, and concentration \cite{Afzal2009}, which makes emotion perception even more difficult to be modeled by a computational model.
	
	Another complexity on the perception side of emotion recognition is the presence of multimodal stimuli in real-world scenarios. The observation of different multimodal characteristics, such as facial expressions, body movement and the way your voice sounds when you emphasized by an affective state is shown to be very important in the determination of perceived emotions \cite{Kret2013}. Until recently, most of the computational models dealing with emotion perception did not take in consideration multimodal stimuli \cite{Soleymani2017}. This is changing gradually given the high capacity of computers to process multimedia information and the advent of techniques such as deep and recurrent neural networks that can learn how to represent such high-level stimuli without the necessity of strong pre-defined feature extractors \cite{Cambria2016}.
	
	The modeling of emotional behavior was also a focus in several studies in the past decade \cite{Pantic2007,Xiao2016}. Most of proposed models in these studies are applied on virtual agents with a strong inclination to use prior knowledge and are mostly developed as rule-based systems. Such systems, although modeling some aspects of emotional behavior, are hardly adaptable and would fail to model situations which were not part of their original interaction scenario.
	
	Some works go further and adapt to real-world-like scenarios. One of the most successful is the work of Lim and Okuno \cite{Lim2014}, which presents the MEI framework to model emotional behavior based on audio and visual stimuli. This framework uses the SIRE paradigm \cite{Lim2012}, which transforms the audio and visual stimuli into abstract representations, which are then used to recognize and generate emotional behavior. Their model is fairly simple and efficient in their task: motherese-like interaction. The limitation of the scenario makes it difficult to be adaptable and extended to more complex scenarios as a social interaction. Also, their expression representation is purely hard-coded, and thus, is only able to recognize very few pre-established emotional states.
	
	Another aspect that is missing in most of the emotion appraisal models is the role of affective representation as a modulator. There is evidence showing that emotion perception is a strong modulator to different brain processes \cite{Sebastian2017}. Affective memory modulates perception \cite{Russel1980}, intrinsic motivation modulates action \cite{DeVries2008} and intuition is built based on emotional behavior \cite{Asada2015}, and yet the proposed models for affective computing tend to ignore these functionalities and model perception and emotion independently, while a combined model would be able to provide a robust framework for a natural perception of emotions and modeling of affective modulation, for example.
	
	\section{Crossmodal Emotion Perception}
	
	We propose an updated version of the Cross-Channel Convolution Neural Network (CCCNN) perception model\cite{Barros2016B}. The CCCNN uses a deep neural network with an integrative cross-channel connection to learn audio-visual representations to describe emotion expressions. It showed to be competitive with state-of-the-art models in the recognition of emotional expressions in the wild, since it learns strong crossmodal representations based on real-world data. Although we do not claim to model neurophysiology of the human brain, the conceptual and topological design of the model are inspired by how the human brain perceives emotional expressions.
	
	
	The use of a multichannel architecture emulates the processing in audio and visual cortices in the primate brain. Although the use of convolution neural networks is not showed to be biologically plausible, the representations learned by the convolution filters seem to show similarity to the representations learned by the primate visual cortex \cite{Dubey2016,Kubilius2016}, and auditory cortex \cite{gucclu2016}. Furthermore, to have unimodal representations learned by specific convolution channels allow us to have a hierarchical behavior on learning features, with the early layers learning feature representations similar to the ones found on early parts of the auditory and visual cortices.
	
	The early individual stimuli layers of the current model learn hierarchical representations of emotion expressions, based on facial expressions and emotional features from speech signals. The deeper layers learn high-level characteristics, such as intonation and facial poses, and thus are used to characterize an expression. In the human brain, the auditory and visual pathway also communicate with each other and are directly connected to other brain regions, such as the superior temporal sulcus (STS). The neurons in the STS encode several different pieces of information including multisensory processing \cite{Senkowski2008} and are directly related to social perception \cite{Beauchamp2015}. The neurons in the STS react strongly for semantic understanding \cite{VanLancker1982}, selective attention \cite{Campbell1990}, emotional experience determination \cite{Grossman2001} and language recognition and learning \cite{Hickok2007}. All these tasks are highly correlated with complex visual and auditory stimuli.
	
	To emulate the visual-auditory pathway, the proposed model makes use of a cross-channel connection in deeper regions of the CCCNN, making sure that unisensory contextual information is actually processed taken into account the two modalities. We maintain the unisensory pathway, but use the cross-channel connections as modulators with feedback connections, similar, in behavior, to different layers of the STS. This topological and conceptual design of our architecture approaches the topological design of emotion circuits for perception in the human brain \cite{Ledoux2000}. 
	
	These updates on the CCCNN architecture introduce an important advantage to the model: the capability to learn unisensory information from crossmodal coincident stimuli. This increases the capability of the model to describe emotion expressions, both in unisensory and crossmodal ways. Also, this makes it possible to train the model using only individual stimuli or crossmodal data, which decreases significantly the restriction on data acquisition during learning. Lastly, this enables the model to describe unisensory and crossmodal stimuli in a similar representation, making possible to apply it in different scenarios.
	
	Using the learned representations to describe crossmodal expression of emotions results in obtaining very robust features which can be used to classify expressions. However, once again following the human behavior during emotion perception, we extend our model to learn to categorize expressions in an online manner. Humans learn to describe expressions very early in childhood \cite{Haviland1987,Morris1998,Schore2015}  and the development of the perceptual mechanisms to describe a face is mature in the teen years \cite{Chaplin2015}. This means that humans learn how to categorize expressions into different emotional concepts, without necessarily continuously learning how to describe a face. That explains the universal emotions concept, as we are able to describe expressions even with different cultural, social and geographical backgrounds. That also explains why persons can adapt and learn how others express their emotions \cite{Glaser2003}.
	
	Our model incorporates the idea of learning to categorize known expressions into different emotional concepts. For that, we propose the use of a recurrent growing-when-required network (GWR), which is trained using the emotion expression representations from the CCCNN as input. This Perception GWR clusters the expression representations into emotional concepts in a dynamic way, learning and forgetting when necessary. By using recurrent connections and context layers, this model can learn temporal characteristics of emotion expressions, and thus, is important for continuous emotion perception. 
	
\begin{figure} 
		\center{\includegraphics[width=0.9\linewidth]{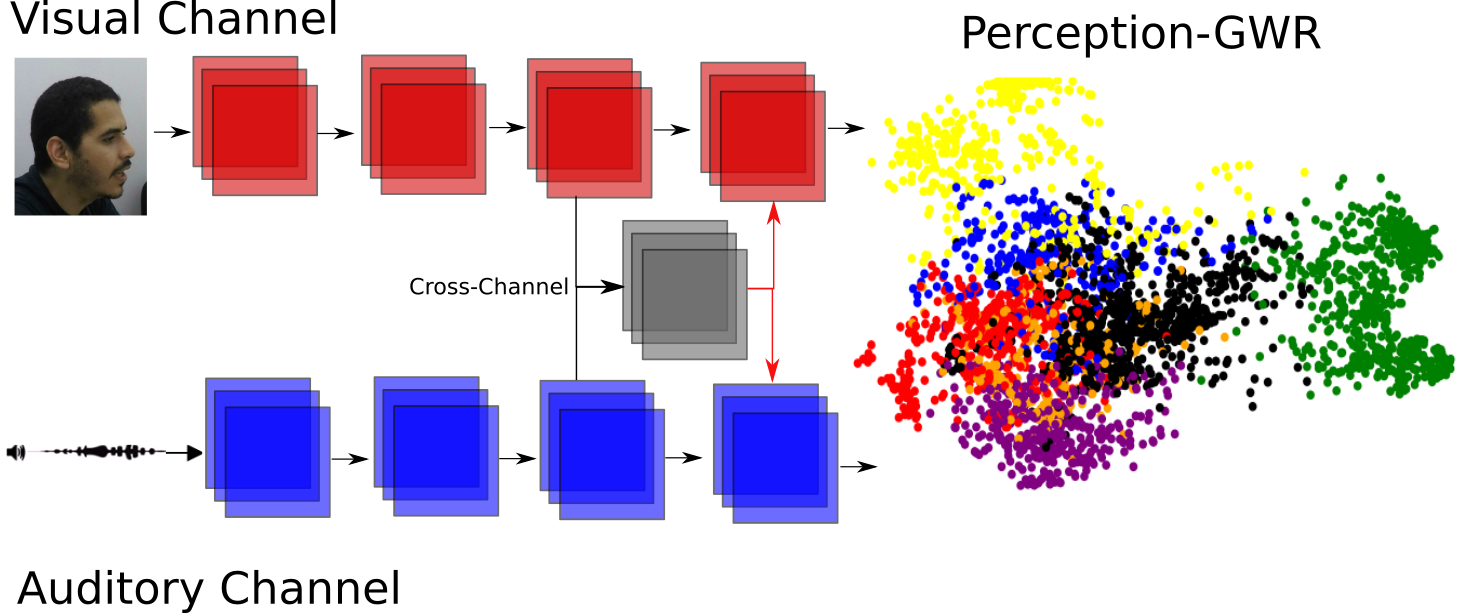}}
		\caption{Cross-Channel Convolution Neural Network with a crossmodal architecture. This topology introduces the use of individual unisensory channels which receive a feedback connection (in red) from a cross-channel layer, that learns how to incorporate crossmodal stimuli. We use the Perception-GWR to categorize learned expressions into emotional concepts in a continuous and self-adaptive manner.}
		\label{fig:topology_CCCNN}
	\end{figure}

	Once the CCCNN is trained and robust emotional feature representations are obtained, the Perception GWR can be initialized to learn with specific or general data. When using the model to describe emotion expressions from persons with a particular cultural background, we can create specific Perception GWRs, or when we want to have a general perception, we can train one general Perception GWR with all the available data. As the GWR is a self-organizing network, we do not need constraints such as labels to train it, meaning that it can be used to learn from several different data sources. Also, we can use the Perception GWR to learn online, having the model clustering emotion expressions while they are presented to it. The topological design of the proposed model is visualized in Figure \ref{fig:topology_CCCNN}.
	
	\subsection{Unisenory Pathways}
	
	
	We can identify if a person is happy or sad by looking only at an image of a facial expression as indicated by Ekman's universal emotions study  \cite{ekman2009}. However, the phenomenon known as micro face expressions can change the categorization of an emotional concept. This phenomenon was first observed by Darwin \cite{darwin1873} and occurs independently of the person self-assessment of the expression, usually in an involuntary way. Ekman demonstrates that facial micro expressions last from 40 to 300 milliseconds, and are composed of involuntary movements or gestures of the components of the face, such as cheeks, eye, and lip movements, and sometimes are not directly related to the expression the person intended to perform. He also shows that microexpressions are too brief to convey an emotion, but usually are signs of concealed behaviors, giving the expression a different meaning. For example, facial micro expressions are usually the way to determine whether someone is angry while using a happy sarcastic expression. In this case, the addition of facial micro expressions as an observable modality can enhance the capability of the model to distinguish spontaneous expressions, but the observation of the facial micro-expression alone does not carry any meaning \cite{Pfister2011}.
	
	To deal with micro and macro face expressions, the visual channel of the proposed model is fed with facial fragments comprising 300ms. We feed a segmented face to the visual channel, which ensures that this channel will learn facial features. In the experiments, the input stream was set to 30 frames per second, which means that the visual channel receives chunks of 9 frames as input. Each of these frames is resized to 128x128 pixels and transformed to grayscale, meaning that the input of the visual channel has a dimension of 9x128x128. To learn spatial and short-temporal representations, we implemented the visual channel with cubic convolutions. In a cubic receptive field, the value of each unit ($	u_{n,c}^{x,y,z}$) at the $n$th filter map in the $c$th layer is defined as: 
	
	\begin{equation}
	u_{n,c}^{x,y,z} = max ( b_{nc} + S_{3}, 0 )
	\end{equation}
	
	\noindent where  $x,y,$ and $z$ represent the index of the unit in each of the dimensions, and  $\max(\cdot,0)$ represents the rectified linear function , which was shown to be more suitable than non-linear functions for training deep neural architectures. As discussed by \cite{Glorot2011}, $b_{cn}$ is the bias for the $n$th filter map of the $c$th layer, and $S_{3}$ is defined as
	
	\begin{equation}
	S_{3} = \sum_{m}\sum_{h=1}^{H}\sum_{w=1}^{W}\sum_{r=1}^{R}w_{(c-1)m}^{hwr}u_{(m-1)}^{(x+h)(y+w)(z+r)},
	\end{equation}
	
	\noindent  $m$ indexes over the set of feature maps in the ($c$-1) layer connected to the current layer $c$. The weight of the connection between the unit ($h$,$w$,$r$) and a receptive field connected to the previous layer ($c-1$) and the filter map $m$ is $w_{(c-1)m}^{hwr}$. $H$ and $W$ are the height and width of the receptive field and $z$ indexes each stimulus; $R$ is the number of stimuli stacked together and represents the new dimension of the receptive field.
	
	The visual channel is built based on the VGG16 \cite{VGG2016} architecture. That means that the proposed network is composed of 10 convolution layers (adapted here as 3D convolutions), but only 4 pooling layers. We use batch normalization within each convolution layer and a dropout function after each pooling layer to give the network stability during the learning process. As in our previous CCCNN architecture, we apply shunting inhibitory fields \cite{Fregnac2003} in our last layer.  Each shunting neuron $S_{nc}^{xy}$ at the position ($x$,$y$) of the $n^{th}$ receptive field in the $c^{th}$ layer is activated as:
	
	\begin{equation}
	S_{nc}^{xy} = \frac{u_{nc}^{xy}}{a_{nc} + I_{nc}^{xy}}
	\end{equation}
	
	\noindent where $u_{nc}^{xy}$ is the activation of the common unit in the same position and $I_{nc}^{xy}$ is the activation of the inhibitory neuron. The weights of each inhibitory neuron are trained with backpropagation. A passive decay term, $a_{nc}$, is a defined parameter and it is the same for the whole shunting inhibitory field. This allows the network to learn very generally emotional representations which will be detailed in the experimental sections. 
	
	The auditory channel receives as an input Mel-Frequency Cepstral Coefficients (MFCCs representation). To extract the MFCCs, a cosine transformation is applied and this projects each value of the Y-axis into the Mel frequency space, which may not preserve locality. Because of the topological nature of 2D filters, the network will try to learn patterns in adjacent regions, which are not represented adjacently in the Mel frequency domain. Abdel-Hamid et al. \cite{AbdelHamid2014} propose the use of 1D filters to solve this problem. The convolution process is the same, but the network applies 1D filters on each value of the Y-axis of the spectrum. That means that the activation of each unit $u_{n,c}^{x}$ at ($x$) position of the $n$th feature map in the  $c$th layer is given by
	
	\begin{equation}
	u_{n,c}^{x} = max\left ( b_{nc} + S_1, 0   \right ), 
	\end{equation}
	
	\noindent where $\max(\cdot,0)$ represents the rectified linear function, $b_{nc}$ is the bias for the $n$th feature map of the $c$th layer and $S_1$ is defined as:
	
	\begin{equation}
	S_1 = \sum_{m=1}^{M} \sum_{w=1}^{W}w_{(c-1)m}^{w}u_{(c-1)m}^{(x+w)},
	\end{equation}
	
	\noindent where $m$ indexes over the set of filters $M$ in the current layer, $c$, which is connected to the input stimuli on the previous layer ($c$-1). The weight of the connection between the unit $u_{n,c}^{x}$ and the receptive field with position $W$ of the previous layer $c-1$ is $w_{(c-1)m}^{w}$.
	
	This means that the filters will learn how to correlate the representation per axis and not within neighbors. Pooling is also applied in one dimension, always keeping the same topological structure.
	
\begin{figure} 
		\center{\includegraphics[width=1\linewidth]{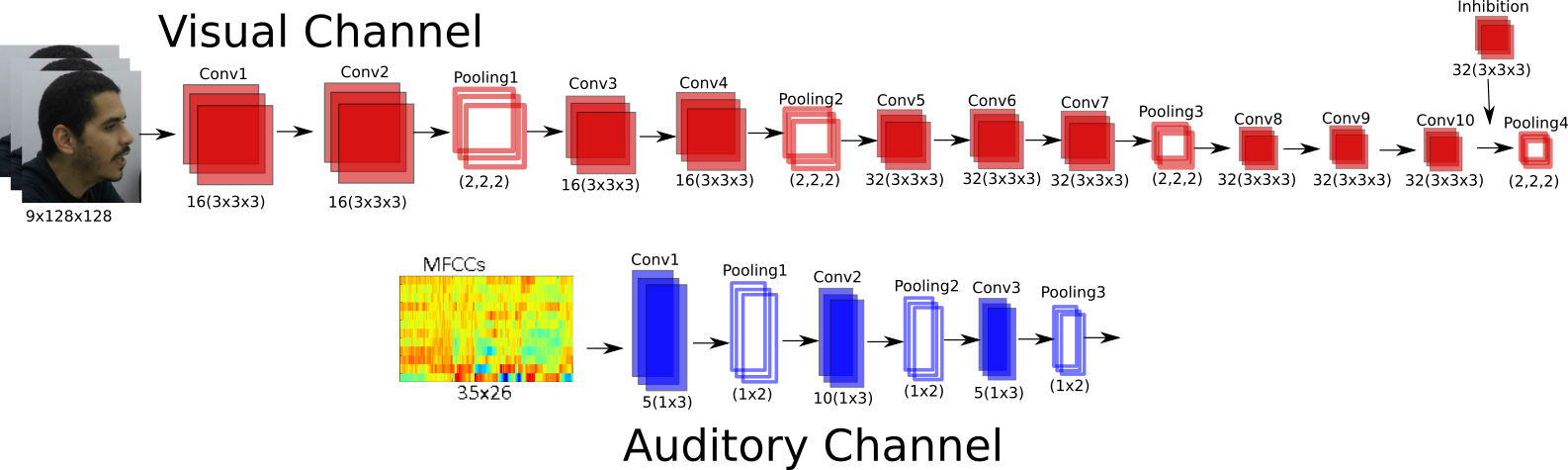}}
		\caption{Detailed architecture and parameters of the Visual and Auditory channels of our CCCNN architecture.}
		\label{fig:individualCCCNNs}
	\end{figure}

	We use audio clips with 1s as input, and each clip is re-sampled to 16000 Hz. We compute the MFCCs of the audio clip and feed it to the network. It is computed over a window of 25ms with a slide of 10ms. We use frequency resolution of 1024, which generated a representation with 35 bins, each one with 26 descriptors. The auditory channel is composed of three layers, each one with one-dimensional filters and followed by a pooling layer. Figure \ref{fig:individualCCCNNs} illustrates the detailed topology of both channels, with precise parameters for each layer.

	As input for the model, we use an input stream with 1s duration. The auditory channel receives the full 1 second as input, but the visual channel receives only 300ms. To synchronize that, for each 1s of input, we randomly cut 300ms of it to feed the visual channel. That guarantees that the visual channel does not learn only the beginning or the end of the input stimuli, but the apex of the expression and during the crossmodal integration, the model learns to associate the visual features with different parts of the audio stimuli.
	
	\subsection{Crossmodal Integration and Feedback}
	
	Computational models for crossmodal learning are usually based on fusion of features and/or decisions at early stage \cite{Wei2010} or at a late-stage \cite{Liu2016,deBoer2016} of processing. Although successful on the tasks of classification or prediction, these models typically rely on individual and independent representations of each stimulus. That is a problem when trying to represent one individual modality again, as the crossmodal representation is basically a new collection of features \cite{Poria2017,Chen2017}. There is a neurophysiological evidence showing that while unisensory information is processed in the human brain, other regions are activated and modulate this processing \cite{Kayser2015}. Having a computational system that takes into consideration unisensory modulation for crossmodal learning would give such system an advantage on both learning unimodal and crossmodal representations.
	
	To emulate this behavior in the proposed model, a modification to the cross-channel architecture is introduced: It uses the crossmodal learning representation as modulator to the deeper layer features of the individual stimuli channels. This is done by using the  layer preceding the last one of each individual channel as an input to the cross-channel architecture and feeding the output of the cross-channel layer to the last layer of the individual channels.
	
	Each of the individual channels have different output dimensions, caused by the different size of the input signals, the number and type of filters and the nature of the data, therefore, we use a normalization technique based on fully-connected layers. Each of the outputs of the pre-last layer of the individual channel is flattened and fed to a fully-connected layer with 100 units which are regularized with an L2 Norm during the weight update $w_{t+1}$:
	
	\begin{equation}
	w_{t+1} = w_{t} - \eta\frac{\partial E}{\partial w_t} + \lambda\sum_{k}^{i}w^2_t
	\label{eq:L2Regularization}
	\end{equation}
	
	\noindent where $w_{t}$ represents the previous weights, $ \lambda$ represents a parameter which controls the relative importance of the regularization term and $\partial E$ represents the error back propagated to this layer. We use a rectified linear unit as an activation function.
	
\begin{figure} 
		\center{\includegraphics[width=0.4\linewidth]{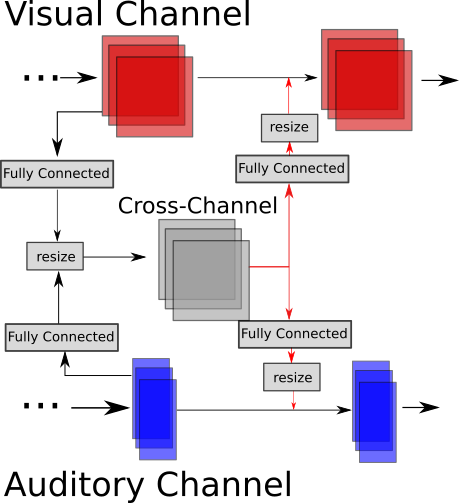}}
		\caption{Detailed topological design of the new cross-channel architecture. The feedback connections are drawn in red.}
		\label{fig:Cross-channelSchematics}
	\end{figure}

	The next step is to resize the output of this layer to a matrix with a dimension of 10x10 and feed both of them, the visual and auditory layers, to the cross-channel architecture. That means that the input for the cross-channel architecture has a dimension of $2x10x10$. Figure \ref{fig:Cross-channelSchematics} illustrates this solution.
	
	The cross-channel is composed of one convolution layer with 8 filters, each one with a size of $3x3$. We used a padding convolution to avoid the reduction of the stimuli dimension. The convolution filters learn how to integrate patterns from both auditory and visual stimuli, and the output of this layer represents an integrated stimuli. We then use this integrated stimulus as feedback for the individual channels. To do that, we again make use of a normalizing fully-connected layer, with the same structure as the one used previously, and a resize operation to allow the output of the cross-channel to have the same dimension as the input of the last convolutional layer of each individual channel. We then use a summing operation and merge the output of the cross-channel and the input of the last convolution layer for each channel. To control the impact of the cross-channel on the individual stimulus channel we use a modulation function when summing the stimuli, described as $M$:
	
	\begin{equation}
	\begin{aligned}
	M = (Cc\gamma) + C_{x}
	\end{aligned}
	\end{equation}
	
	where $Cc$ represents the resized output of the cross-channel and $C_{x}$ represents the original input of the pre-last convolution layer of each individual channel. The modulation factor is defined as $\gamma $ and it has a value within the interval of $[0,1]$. This factor modulates the contribution of the cross-channel on the unisensory representation ($0$ to have no contribution and $1$ to have a large contribution). 
	
	By using the cross-channel to generate a modulator for the unisensory input, we are able to maintain an individual representation of each stimulus, which is important for representation of the emotion expression by one sensory modality. However, the individual representation is affected by crossmodal coincident stimuli, meaning that a facial expression which is present when the person is crying will most likely be merged. The network will learn how to represent both stimuli individually, but with a contribution from the other modality which can be controlled by the modulating factor $\gamma$.
	
	After the network is trained, the architecture can also be used as an unisensory descriptor if the crossmodal modulator is set to zero. This modeling decision is made because the network does not need to have both modalities to generate a representation, in fact, it can generate representations of expressions from both modalities individually or jointly, which makes it extremely flexible.

	\subsection{Clustering of Emotional Concept}
	
	To classify emotion expressions is a difficult task: First the observation of various different modalities is necessary. Second, the concept of emotion itself is not precise, and the idea of classifying what another person is expressing based on very strict rules makes the analysis of such models difficult.  
	
	Dealing with a set of restricted emotions as defined by Ekman \cite{Ekman1984} is a serious constraint to affective and social systems. Humans have the capability to learn how to describe emotional expressions, and then adapt their internal representation to a newly perceived emotion from a different person. This is explained by Hamlin \cite{Hamlin2013} as a developmental learning process. Her work shows that human babies perceive interactions into two very clear directions: positive and negative. When the baby is growing, this perception is shaped based on the observation of human interaction. Eventually, concepts such as the six universal emotions are formed. After observing individual actions toward others, humans can learn how to categorize complex emotions which is a precursor for understanding concepts such as trust and empathy \cite{harter1987,Lewis2012,Pons2004}.
	
	To create a model of a developmental emotion perception mechanism, we focus on the dimensional model of affect \cite{Russel1980} . Following the findings of Hamlin \cite{Hamlin2013} on developmental learning, we train a CCCNN to learn strong and reliable emotion expression representations in different modalities. We use the CCCNN architecture to create expression representations, we then attach a fully-connected layer to the individual CCCNN outputs, to create a joint representation which will be fed to a recurrent Growing-When-Required (GWR) layer \cite{Parisi2017} to learn emotion concepts. As stated before, the GWR is an unsupervised learning architecture which has the ability to grow, by adding more neurons, in any direction. This means that the network is not restricted to a number of neurons, either by any topological structure, and will relearn and update the number of neurons and the structure to represent the changing data structure. In our GWR each neuron has a set of weights $\textbf{w}_j$ and $K$ contexts $\textbf{c}^k_j$ (with $\textbf{w}_j,\textbf{c}^k_j\in\mathbb{R}^n$). That means that each neuron has recurrent connections to its own contexts, and will encode a prototype expression which contains sequential information of the input stimuli. As the input stimulus is of an emotion, each neuron in our GWR will encode a prototype expression represented by a certain sequence. 
	$N$
	Once we encode an expression with the CCCNN( $\textbf{x}(t)\in\mathbb{R}^n$), we can find the best-matching unit (BMU) $\textbf{w}_b$ in our recurrent GWR  with $N$ neurons using the following computation:
	
	\begin{equation} \label{eq:GetB}
	b = \arg\min_{j\in N} \left( \alpha_0 \Vert \textbf{x}(t) - \textbf{w}_j  \Vert^2 + \sum_{k=1}^{K}\ \alpha_k \Vert \textbf{C}_k(t)-\textbf{c}_{k,j}\Vert^2 \right),
	\end{equation}
	\begin{equation}\label{eq:MergeStep}
	\textbf{C}_{k}(t) = \beta \cdot \textbf{w}_{I-1}+(1-\beta) \cdot \textbf{c}_{k-1,I-1},
	\end{equation}
	
	where $\alpha_i$ and $\beta \in (0; 1)$ are parameters which modulate the impact of the perceived expression on the contextual information which was perceived previously, $\textbf{w}_{I-1}$ is the weight of the BMU at $t-1$, and $\textbf{C}_{k}\in\mathbb{R}^n$ is the global context of the network with $\textbf{C}_{k}(t_0)=0$.
	
	The GWR grows to adapt to the input data, meaning that the expression distribution which is shown to the network is actually better fitted, which produces a more robust learned representation. For that, the network inserts a new neuron when the network activity $a(t)$ of the habituated neuron is smaller than the insertion threshold $a_T$: a new neuron $r$ is created if $a(t)<a_T$ and $h_b<h_T$.
	

	The GWR algorithm gives our model three important characteristics: it provides the capability to detect novelty, to adapt new expressions in the moment they are presented to the network, it allows modeling of temporal aspects of emotion perception and, most importantly, it has the capability to learn and forget concepts. That means that we can use our GWR to learn how to associate different expression modalities, identify and learn not previously seen expressions, cluster them into new emotional concepts, and forget concepts which are not important anymore.

	\begin{figure} 
		\center{\includegraphics[width=0.9\linewidth]{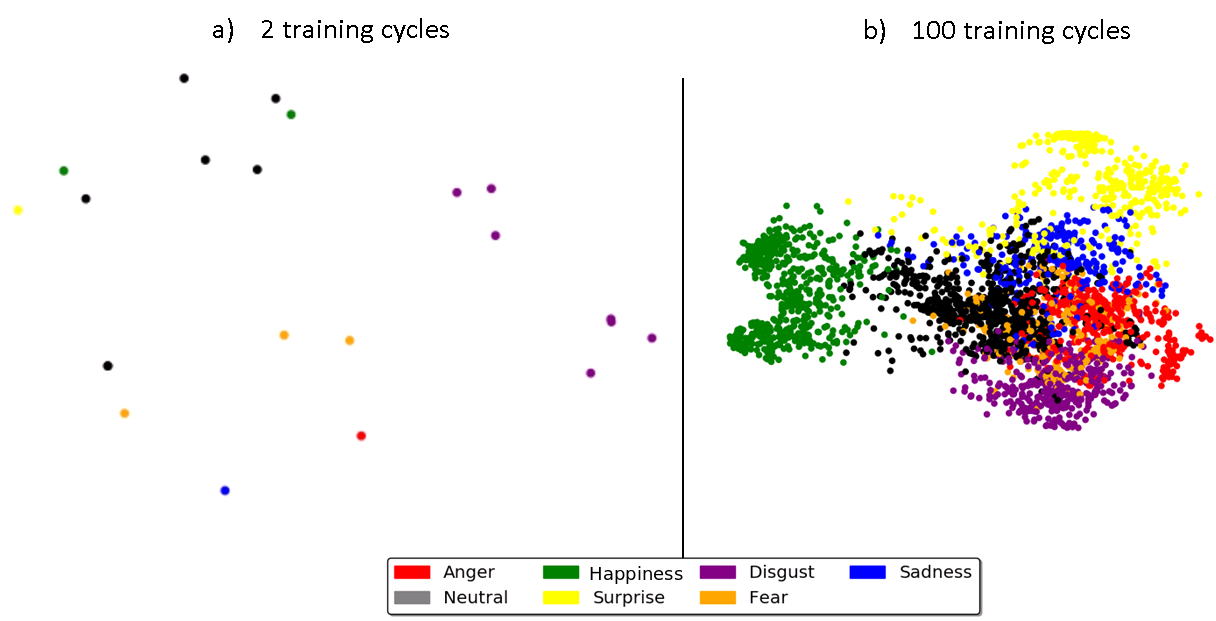}}
		\caption[Illustration of our general Perception GWR.]{We proceed to train a Perception GWR, which will maintain our entire representation of multimodal emotion expression perception. The figure illustrates the general network trained with emotion expressions from all our corpora, in the first training cycle on the left, and after 100 ones on the right.}
		\label{fig:GeneralGWR}
	\end{figure}
	
	We use a GWR model to learn general multimodal emotion expressions, which we named the Perception GWR. This model represents the general knowledge of our perception architecture and is able to identify several different types of expression. We train this Perception GWR with different expressions coming from all our corpora, in a way that it produces the most general representation possible. Figure \ref{fig:GeneralGWR} illustrates the overall proposed network in the first interaction, on the left, and in the last interaction, on the right. In this example, we use a categorical classifier to assign each neuron of the Perception GWR one of the universal emotions after we trained the network just for visualization purposes.


	

	
	\section{Modeling Intrinsic Emotion Concepts}
	
	In a neural network, memory is usually related to the weights and how they create a separation plane withing the input stimuli in order to represent them \cite{OReilly2006}. In self-organizing architectures a different type of memory is encoded in the networks: instead of carrying representation about how to categorize the input stimuli , the weights of the neurons in such model encode a prototype of the input data. In a self-organizing learning, each neuron can be seen as a memory unit, which is trained to resemble the input data \cite{Konohen1998}. 
	
	One common use of self-organizing neural networks is as associative memory tasks \cite{Konohen1987,Konohen2012}. In such tasks, the neurons of a self-organizing network will learn how to memorize associations between two concepts. We exploit this feature in the self-organizing layer of our CCCNN to associate auditory and visual modalities, and then generate a memory of what the network learned, grouping similarly learned concepts together.
	
	Training our recurrent GWR with different expressions gives us a very powerful associative tool which will adapt to the expressions which are presented to it. By adapting the learning and forgetting parameters of the GWR we can determine how long the network will keep the learned information, simulating different stages of the human memory process. For example, training a GWR to forget quickly will make it associate and learn local expressions and by decreasing the forgetting factor it is possible to make the network learn more expressions, meaning that it can adapt its own topology to a set of expressions that was presented in a mid- to long-time span. 
	
	Using the GWR we can create different types of emotional memory. By having multiple GWRs, with different learning and forgetting factors, we can simulate several types of emotional memory: short- and long-term memory, but also personalized affective memory, related to a scene, person or object, and even mood. By feeding each of this memories with the Best-Matching Unit (BMU) of the Perception GWR, we can create an end-to-end memory model, which will learn and adapt itself based on what was perceived. The Perception GWR can learn new expressions if presented, and each of the specific memories will adapt to it in an unsupervised fashion.  In this paper, we introduce two of these concepts: the Affective Memory and the Mood.
	
	\subsection{Affective Memory}
	
	We implemented the Affective Memory as a specific GWR trained for one specific person. That means that for each person the network has a separated Affective Memory. This is an important feature for social robots as the way a person is behaving while in contact with a robot can modulate the behavior of the robot's response. This specific GWR is trained in an online fashion with a low forgetting rate, which means that it will learn and evolve over time based on how that specific person behaves during the interaction with a social robot with embedded GWR.

	\begin{figure} 
		\center{\includegraphics[width=0.9\linewidth]{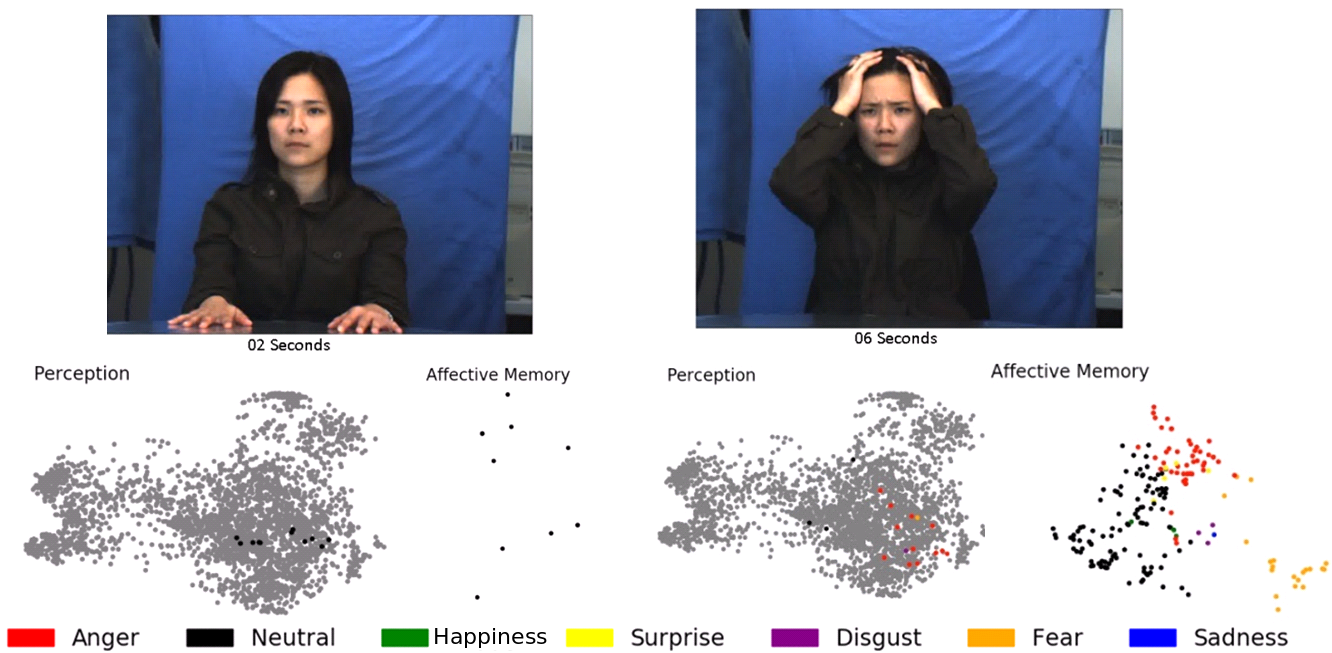}}
		\caption[Illustration of our Affective Memory GWR.]{Using the expressions depicted on the Perception GWR, we proceed to train an Affective Memory GWR for a video. The network on the left illustrates the Affective Memory on the start of the video (02 seconds) and on the right at the end of the video (06 seconds). The colored dots in the Perception GWR indicate which neurons were activated when the expression is presented and the emotion concept associated with them. The colored neurons on the Affective Memory indicate which emotion concepts these neurons code. }
		\label{fig:EmotionExampleGWRNetwork}
	\end{figure}
	
	Figure \ref{fig:EmotionExampleGWRNetwork} illustrates a GWR architecture used to represent an Affective Memory for a video sequence. We first use the Perception GWR to detect which expressions were performed, and we feed this information to the Affective Memory. At the beginning, (represented by the topology on the left in Figure \ref{fig:EmotionExampleGWRNetwork}), it is possible to see that the network memorized mostly neutral expressions. In the end of the training, different concepts emerge while the person is expressing different emotions. By changing the insertion threshold of this network, we can let it learn the expressions from the whole video or just in one part of it.  In our experiments, the Affective Memory network was trained using a insertion threshold of $0.01$. 
	
	The Affective Memory can be used both as a bias of how a person usually behaves to other agents, but also as an independent emotional behavior descriptor. This module presents an individual memory of how the person is behaving over time, and thus, could be used as a complex emotional behavior descriptor.

	\subsection{Intrinsic Mood}
	
	The mood is usually described as an intrinsic emotional state which is modulated by different sensory and physiological factors \cite{Ketai1975,Batson1992} and modulates decision, motivation, and actions \cite{rottenberg2005,Cardi2015}. In computational models, the mood is usually related as an intrinsic state which modulates motivation to solve tasks \cite{Abro2016} or as a part of empathy models \cite{Paiva2017}. In our model, the Mood layer is an intrinsic concept of an emotional state which is modulated by perception, however not imitating nor behaving as an empathy model. One important aspect of our Mood layer is that it does not discriminate the subject, which is different from our Affective Memory layer which is subject dependent. That allows us to use our Mood layer to construct an emotional state which can be used to describe the general emotion behavior towards a robot over a specific period of time with different subjects in the interaction. That means that our Mood layer evolves accordingly to what was perceived, but not with a linear or direct relation. To achieve that, we model mood with a GWR with a higher insertion threshold of $0.001$, which is smaller than the Affective Memory GWR, which means that our Mood layer presents more updates and has a less stable content. That is important as the Affective Memory layer must be more stable  as it models a user dependent behavior which lasts longer than a mood.
	
	One important difference between the mood and the Affective Memory GWR is that the later works as a memory model, receiving as input the BMUs from the Perception GWR and modeling what was perceived over a certain period of time. The mood network receives as an input the arousal/valence descriptors from the BMUs of both the Perception GWR and Affective Memory. That makes the mood an independent module which describes an intrinsic emotional state on an arousal and valence dimension. 
	
	The neurons in the mood GWR store the emotional concept of what was perceived in the format of arousal/valence, and not prototypes of the expression itself as the Perception GWR. That represents the highest level of abstraction in our framework, and thus, the final affective encoding. By learning arousal and valence distributions, the Mood is an independent module which does not remember what was perceive exactly, but it models the emotional representation that the expression caused during the interaction. 
	
	The Affective Memory impacts the mood by modulating what was perceived. As humans, we have an emotional bias towards persons that we already know. We can be excited to meet a friend and this excitement can impact on how we perceive the world around us. To model that, we propose here a modulation from the Affective Memory to the Mood. 
	
	The modulation is applied as a function and calculates the impact of the Affective Memory into the Mood. First, we have to identify the current Affective Memory state, based on the mean of the valences of all the neurons of the Affective Memory. Then, we calculate the modulator factor $M$:
	\begin{equation}
	M = \left\{\begin{matrix}
	v_p > 0.5, & e+e.\left(\frac{1}{e^-{v_m}} \right ) \\ 
	v_p = 0.5,&  e\\ 
	v_p < 0.5, & e-e.\left(\frac{1}{e^-{v_m}} \right ) 
	\end{matrix}\right.
	\end{equation}
	
	\noindent where $v_p$ is the valence of the perceived expression, $e$ is a constant indicating the modulator strength, and $v_m$ is the mean valence of the memory. The modulator factor indicates the strength of the relation between the perceived expression and the Affective Memory. It will increase if the valences of the perceived expressions and memory are similar, and decrease if not. This follows findings that describe the change of the mood based on empathy with different persons \cite{nezlek2001}. When knowing an empathic person, the impact that this interaction have on the person's own mood is more positive than when interacting a person which is known to be unemphaticc.
	
	We then use the modulator factor $M$ to enforce the arousal and valence stimuli coming from the Perception GWR to the Mood by repeating $M$ times the stimuli to update the mood.

	\section{The Emotional Deep Neural Circuitry}
	
\begin{figure} 
		\center{\includegraphics[width=0.9\linewidth]{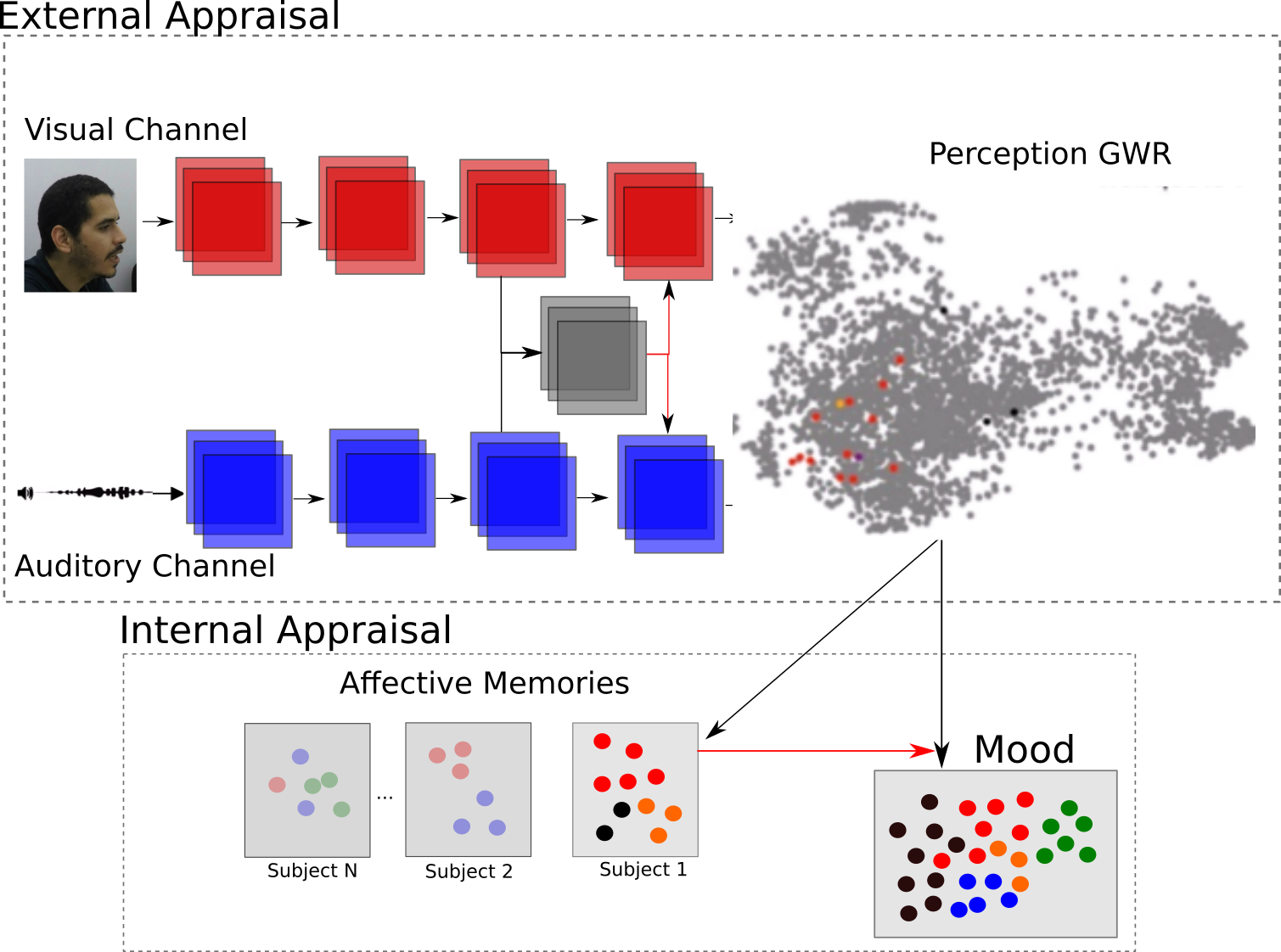}}
		\caption[Illustration of the proposed Emotional (deep) Neural Circuitry.]{The Emotion Deep Neural Circuitry which integrates internal and external appraisal layers. The red arrows represent the modulation connections applied by the cross-channel and the Affective Memory layers.}
		\label{fig:emotionDeepNeuralCircuitry}
	\end{figure}

	To integrate the systems and methods we proposed, the Emotional Deep Neural Circuitry, as illustrated in Figure \ref{fig:emotionDeepNeuralCircuitry} is introduced. This framework integrates the proposed CCCNN with the Perception GWR, the Affective Memory, and the Mood layers. With this framework, it is possible to represent crossmodal emotion expressions in a continuous/online manner and cluster the expression into different emotional concepts. 
	
	For each participant, a new Affective Memory model is created and only updated when that particular participant is present in a scene. This creates an individual measure for each participant, which gives us a tool to measure his/her expressive interaction with the robot in a certain time-span.  Such memory could be related to the concept of affection and empathy, as it will store how a particular subject behaved while interacting with the robot. The Mood is directly modulated by the Perception GWR and Affective Memory and encodes the robot's own perception based on a certain range of past interactions. That means that if the robot is in a negative mood and interacts with a person which it relates positive expressions to in the past, the chances of the robot to change its mood towards a positive one is higher. 
	
	The Emotional Deep Neural Circuitry is trained in several steps. The CCCNN is trained with pertinent data, to achieve a very robust crossmodal representation of the emotional expression. Without this strong pre-training, our model becomes weakly reliable, as all the other models depend directly on the expressions representations. This mechanism demands more time for training, as its implemented with a deep neural network and requires a large amount of data to learn meaningful representations. 
	
	The Perception GWR is pre-trained with a large number of emotional expressions. Due to the characteristics of the CCCNN, we can train the Perception GWR with unisensory or crossmodal data, which increases the variety of the learned clusters. That means that the Perception GWR can learn new expressions and emotional concepts, which were not present during the CCCNN training. Lastly, our Affective Memory and Mood are trained continuously, while performing the experiments. This way, each of these mechanisms can learn different information from the interactions.
	
	To estimate the valence, used for the memory modulator, we use a Multi-Layer Perceptron (MLP). We train this arousal and valence MLP using the Perception GWR BMU as input, meaning that this network can be trained also using unimodal or crossmodal data. Similarly, we use another MLP to classify the categorical emotions, which help us to identify clusters in the networks and for visualization purposes.

	\section{Evaluating the Model}
	
	To evaluate the model, we propose a series of experiments. Each of these experiments was designed to evaluate a specific module or different properties of the proposed framework.
	In this paper, we evaluate two aspects of our model: the external and internal appraisal modules. First, we evaluate the capability of the model in describing episodic emotion expressions, related to the internal appraisal modules and emotion perception. Our second evaluation demonstrates how our model uses the internal appraisal module, composed of the mood GWR and different affective memories to describe long-term emotion behavior.
	
	To evaluate the perception modules, we make use of three datasets with different sensory modalities: The Berlin Emotional Speech Database (EmoDB) \cite{Burkhardt2005} corpus is used to train and evaluate the auditory channel, the Face Expression Recognition Plus dataset (FER+) \cite{Barsoum2016} corpus is used for the visual channel and The One-Minute Gradual-Emotional Behavior dataset (OMG-Emotion dataset) \cite{Barros2018OMG} is used for the cross-channel evaluation and the emotional concept clustering. 
	
	To evaluate the intrinsic appraisal capabilities we propose the novel KT-Emotion Interaction corpus. The corpora we selected for evaluating our framework represent the state-of-the-art on emotion expression recognition. They are among the most challenging corpora in this field and provide a very close relation to a real-world scenario. However, none of the available corpora actually is suited to the formal evaluation of the entire model. This mostly is caused by the lack of a measurable and reproducible interaction on these datasets, where we could evaluate different emotional behavior over controllable factors. The KT Emotion Interaction corpus was designed to leverage studies on emotion appraisal in real-world human-robot interaction scenarios. By merging a controlled environment with a script-less interaction, we are able to have a close-to-real scenario where we can evaluate long-term emotion appraisal models.

	\subsection{The Berlin Emotional Speech Database (EmoDB)}

	The Berlin Emotional Speech Database (EmoDB) \cite{Burkhardt2005} is an auditory emotion corpus with data spoken in the German language. This was collected from five different actors speaking 10 different text sentences with 7 emotional intonations: Anger, Boredom, Disgust, Anxiety/Fear, Happiness, Sadness and Neutral. The dataset has a total of 500 utterances with different lengths, varying from 1s to 5s of duration. To make it standard for our model, we cut each of these utterances into 1s pieces which gives us a total of 1211 utterances. Although this dataset is in German, the principle of universal emotions of Ekman was shown to be present when using this corpus to learn speech emotion features and to generalize them on datasets contain speech in other languages \cite{Xiao2016}.

	\subsection{Face Expression Recognition Plus (FER+)}
	
\begin{figure} 
		\center{\includegraphics[width=0.9\linewidth]{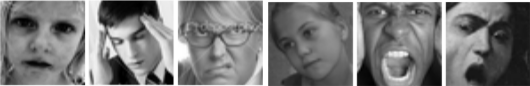}}
		\caption{Examples of different faces on the FER+ dataset.}
		\label{fig:exampleFER+}
	\end{figure}

	The Face Expression Recognition (FER) dataset became very popular in the last years as part of the Facial Expression Recognition Challenge. This dataset has 28.709 face expressions separated into seven categories: Anger, Boredom, Disgust, Anxiety/Fear, Happiness, Sadness, Surprise and Neutral. Although popular, this dataset had some inherent problems on how the images were labeled, as each image had only one final label. As emotion expressions can be perceived very differently, to have a single label per image constrained the capacity of models to generalize well, which was reflected in the relatively low performance with state-of-the-art models \cite{Kahou2016}. To solve this, the FER+ dataset \cite{Barsoum2016} was proposed. This corpus takes all the images from the FER corpus and re-label them using a crowdsource annotation strategy. They collected 10 annotations per image and let them available during training and evaluation of the corpus. This strategy improved the accuracy performance of the model in more than 20\%. Figure \ref{fig:exampleFER+} illustrates examples of data points of the FER+ dataset.
	
	\subsection{The One-Minute Gradual-Emotional Behavior dataset (OMG-Emotion dataset)}
	
\begin{figure} 
		\center{\includegraphics[width=0.8\linewidth]{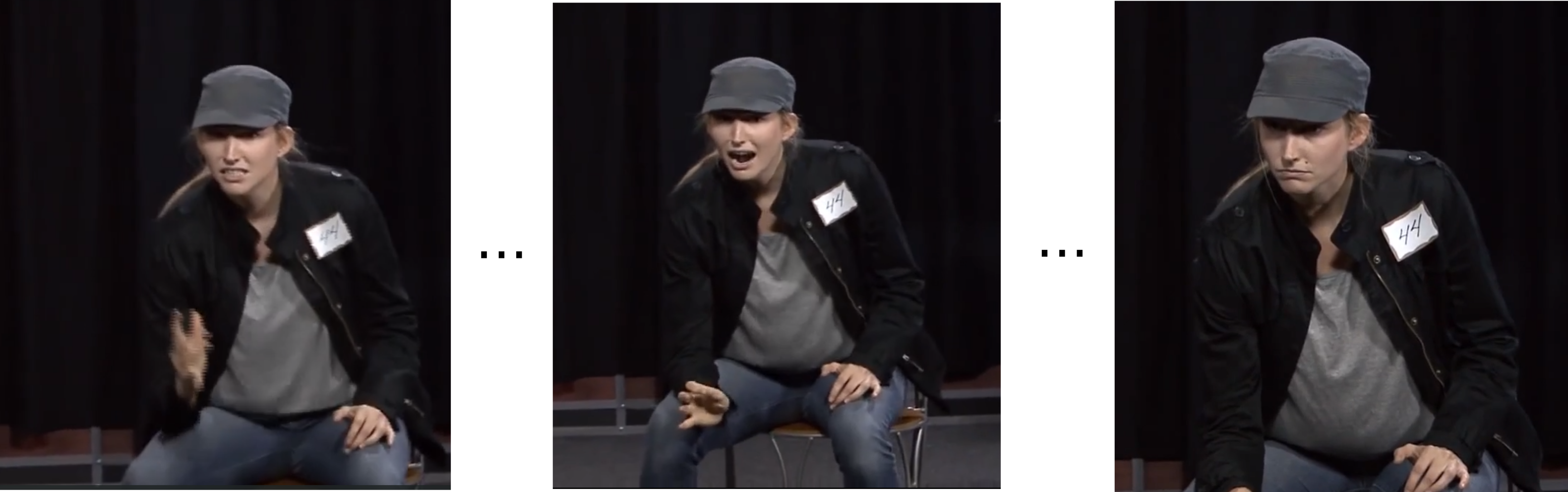}}
		\caption{Examples of different frames of one video from the OMG-Emotion dataset.}
		\label{fig:exampleOMGEmotion}
	\end{figure}

	This dataset is composed of Youtube videos which are about a minute in length and are annotated taking into consideration a continuous emotional behavior. The videos were selected using a crawler technique that uses specific keywords based on long-term emotional behaviors such as "monologues", "auditions", "dialogues" and "emotional scenes". A total of 675 videos were collected, which sums around 10h of data. All the videos were annotated using the Amazon Mechanical Turk platform, and each utterance in each video has at least 5 unique annotations. A total of 4500 utterances were annotated as categorical emotions (Anger, Disgust, Fear, Happiness, Sadness, Surprise and Neutral) and a continuous interval between $[0,1]$ representing arousal and $[-1, 1]$ representing valence. Figure \ref{fig:exampleOMGEmotion} illustrates examples of one video on this dataset.
	
	\subsection{KT-Emotion Interaction Corpus}
	
	To evaluate our model properly, we propose here a new corpus based on human-human and human-robot interaction. To demonstrate how our framework works properly, we need a dataset which has long-term emotion behavior that changes naturally over time. To make comparisons fair and the analysis of our intrinsic emotion concepts formation, these interactions must be controllable but with enough personal individualism, so we can model each person's own behavior. We decided to record this dataset with two interactions: one based on human-human communication and one based on human-robot communication. That also allows us to use our framework to analyze the differences in the nature of human behavior while interacting with another human or a robot.
	
\begin{figure} 
		\center{\includegraphics[width=0.9\linewidth]{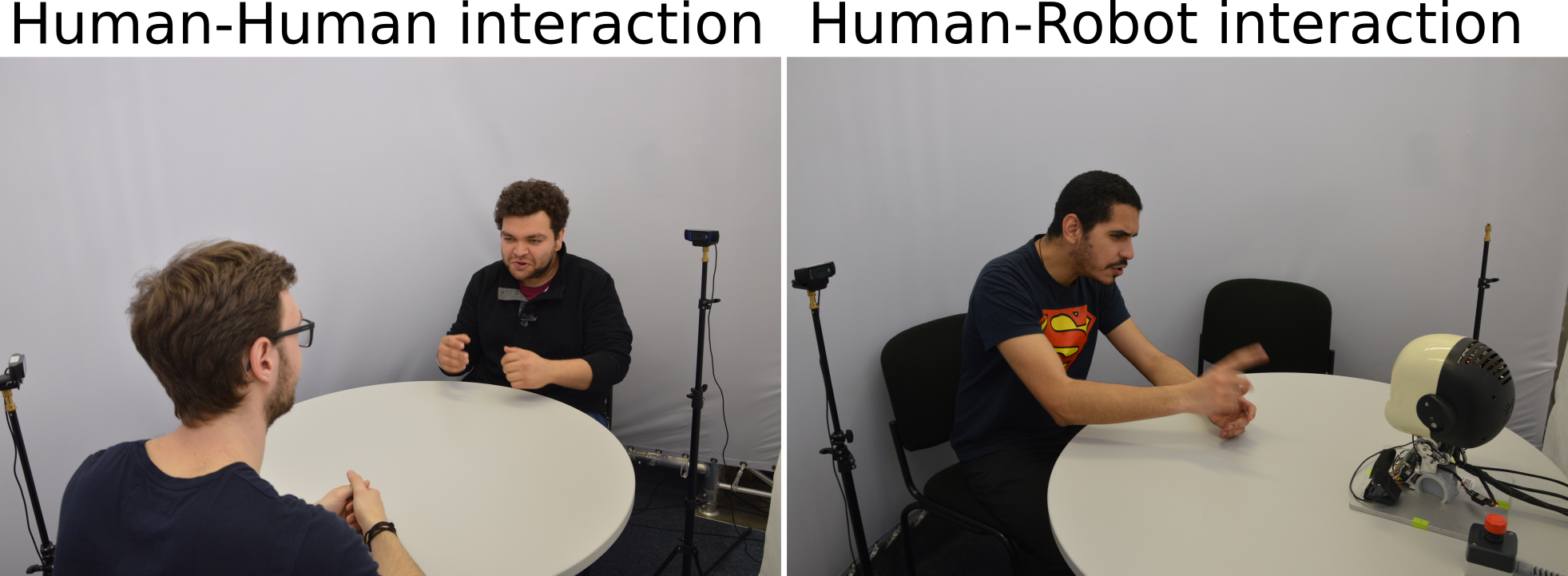}}
		\caption{The two scnearios of the KT-Emotion Interaction corpus: Human-Human interaction and Human-Robot interaction.}
		\label{fig:KTScneario}
	\end{figure}

	The data recording happened in two different scenarios. The first one recorded human-human interactions (HHI), and the second one human-robot interactions (HRI), where a human interacts with an iCub robot \cite{Metta2008}. We annotate the data using dimensional and categorical representations.  In both scenarios, two subjects conducted a fictitious dialogue based on a certain topic. The subjects were seated at a table across from each other. Images from the subjects´ faces and torsos and audio were recorded. In one of the scenarios, both subjects are humans, however, in the second scenario, one of the subjects is replaced by a robot. Figure \ref{fig:KTScneario} illustrates both scenarios.
	
	For both scenarios, we created two roles: an active and a passive subject. Before initiating each dialogue session, we gave to the active subject a topic, and he or she should introduce it during the dialogue. The passive subject was not aware of the topic of the dialogue, and both subjects should improvise. The subjects were free to perform the dialogue as they wanted, with the only restriction of not standing up nor changing places. The following topics were available: 
	
	\begin{itemize}
		\item Lottery: Tell the other person he or she won the lottery.
		\item Food: Introduce to the other person a very disgusting food.
		\item School: Tell the other person that his/her school records are gone.
		\item Pet: Tell the other person that his/her pet died.
		\item Family: Tell the other person that a family member of him/her is in the hospital.
	\end{itemize}
	
	\begin{figure} 
		\center{\includegraphics[width=0.9\linewidth]{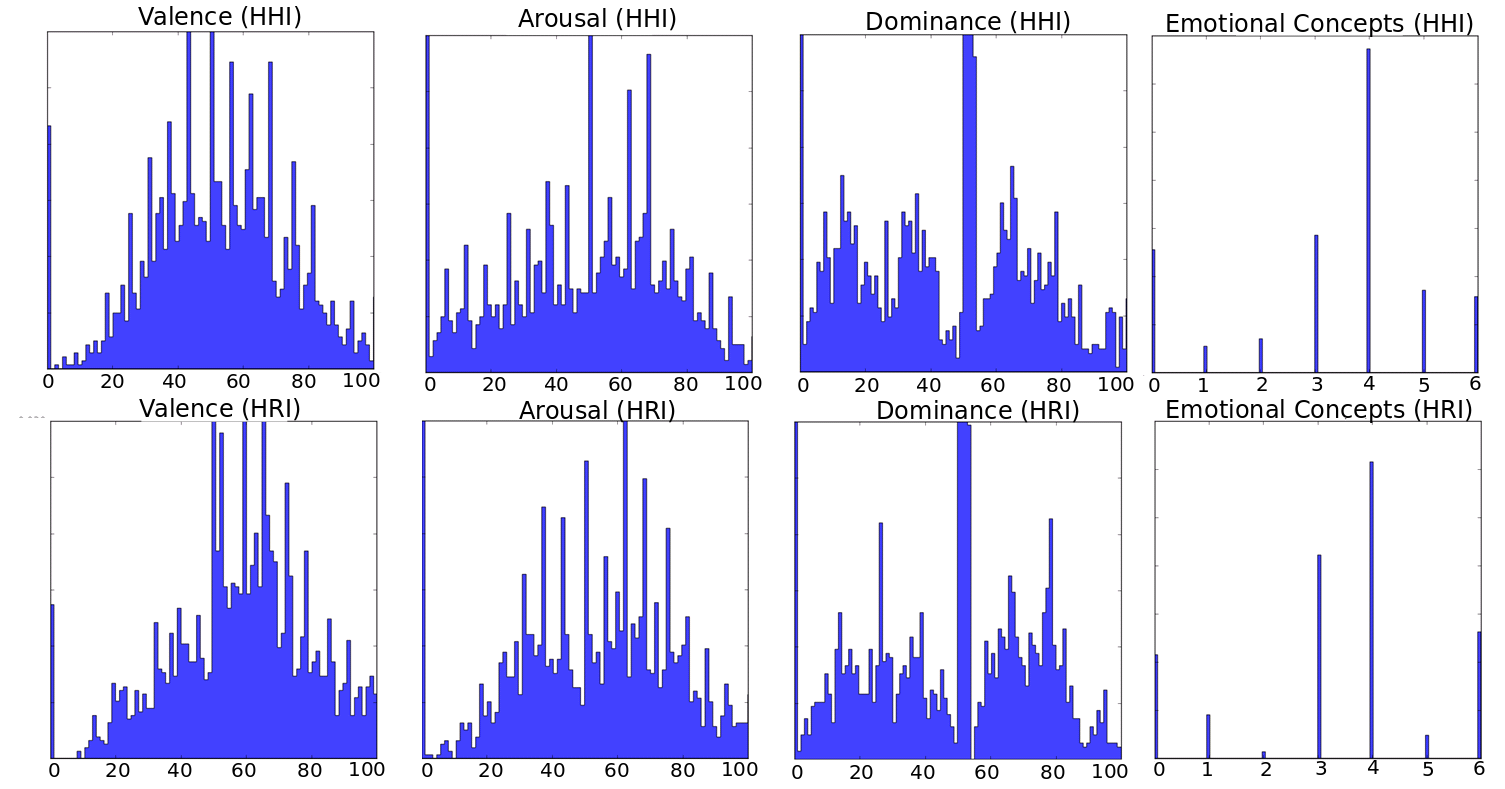}}
		\caption[Analysis of the general data.]{Histogram of the annotations for both scenarios. Int the emotional concepts histogram, the x-axis represents the following emotions: 0 - Anger, 1 -Disgust, 2 - Fear, 3 - Happiness, 4 - Neutral, 5 - Sadness and 6 - Surprise.}
		\label{fig:dataLabelGeneral}
	\end{figure}
	
	These topics were selected in a way to provoke interactions related to at least one of the universal expressions each: ``Happiness", ``Disgust'',``Anger",  ``Fear'', and  ``Sadness". To none of the subjects, any information was given about the nature of the analyses, to not bias their expressions. After each dialogue session, the role of the active subject was given to the previously passive subject and a new topic was assigned. The HHI scenario had a total of 7 sessions, with 14 different subjects, two participating in each session. Each session had 6 dialogues, one per topic and an extra one where the two subjects introduced each other, using a fake name. Each subject only participated in one session, meaning that no subject repeated the experiment.  A total of 84 videos were recorded with a sum of 1h05min of recordings. Each video had a different length, with the longer one having 2 minutes and 8 seconds and the shorter one with 30 seconds.  The HRI scenario had one session more than the HHI, with a total of 9 sessions. In the HRI scenario, each session had 5 dialogues, one per topic, without the introduction dialogue. A total of 45 videos were recorded, summing 2h05min of videos. As happened with the HHI, each video has a different length and the longer one had 3min40s and the shorter one with 1min30s. The subjects were collected from the informatics department of the University of Hamburg and the experiment was approved by the local ethics commission.
	%
	%
	
	To annotate the video material, each video was divided into 10s intervals. Thirty nine different annotators  categorized each sequence into one of the six universal emotions [ reff Ekman]and into arousal, valence, and dominance using a scale in the interval of [0,1]. Figure \ref{fig:dataLabelGeneral} shows the histogram for the valence, arousal, dominance and the emotional labels for all the annotations in both scenarios. It is possible to see that the annotations for all of these dimensions are normally distributed, showing a strong indication that most of the interactions were not so close to the extremes.
	%
	%
	
\begin{figure} 
		\center{\includegraphics[width=1\linewidth]{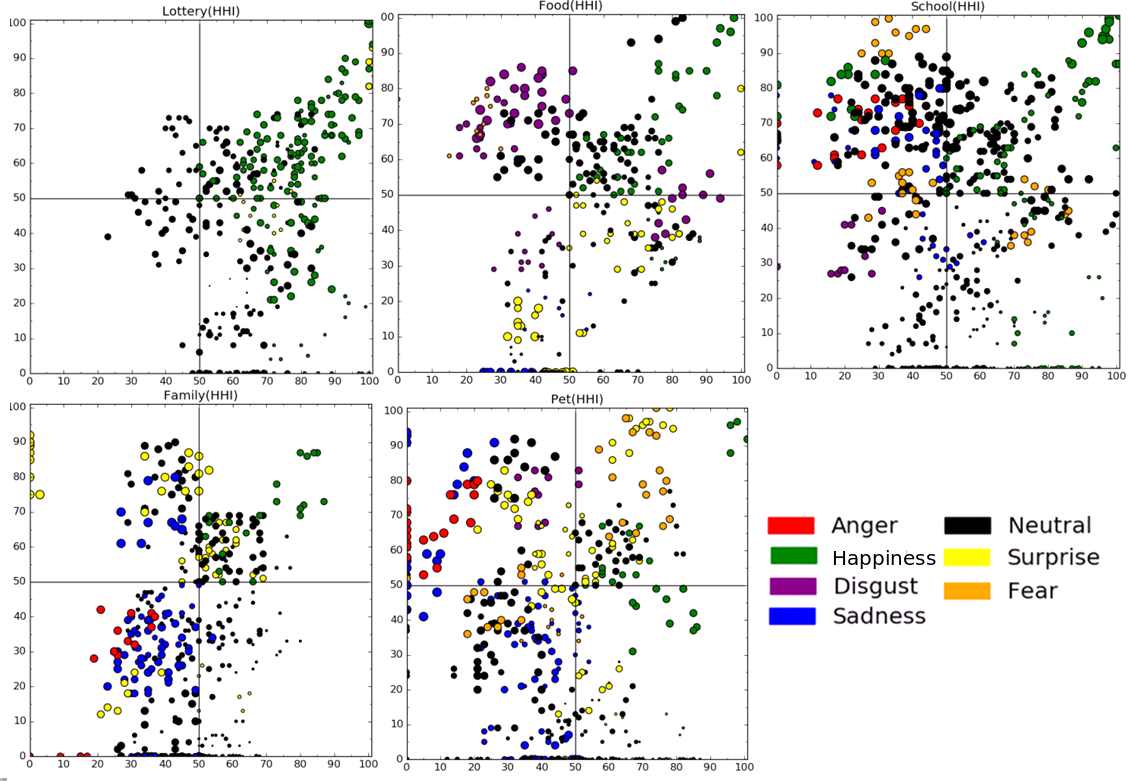}}
		\caption[Analysis on the HHI data per topic.]{Plots that shows the distribution of annotations for the HHI scenario, separated by topics. The x-axis represents valence, and the y-axis represents arousal. The dot size represents dominance, where a small dot is a weak dominance and a large dot a strong dominance.}
		\label{fig:dataLabelTopicHHIAll}
	\end{figure}

\begin{figure} 
		\center{\includegraphics[width=1\linewidth]{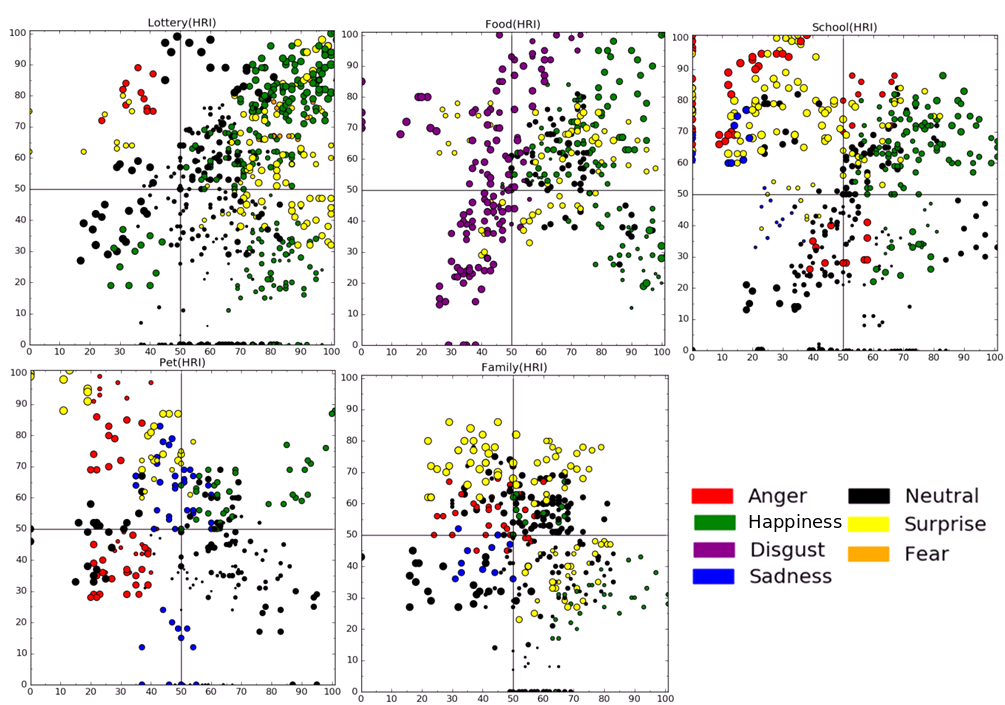}}
		\caption[Analysis on the HRI data per topic.]{Plots that shows the distribution of annotations for the HRI scenario, separated by topics. The x-axis represents valence, and the y-axis represents arousal. The dot size represents dominance, where a small dot is a weak dominance and a large dot with a strong dominance.}
		\label{fig:dataLabelTopicHRIAll}
	\end{figure}

	We provided a distribution plot per topic for each of the scenarios in Figures \ref{fig:dataLabelTopicHHIAll}, for the HHI topic,  and \ref{fig:dataLabelTopicHRIAll}, for the HRI topic. By analyzing the videos of individual topics we provided a knowledge about how different persons reacted while within the same topics. This is an important measure to show how the different topics produce a behavioral pattern even while different persons took part in the interaction. It is possible to see also the differences in the behavior of persons while performing the same topic with another person or with the robot. 
	%
	%
	%
	%
	
	We also calculated the Concordance Correlation Coefficient (CCC) for the annotations of videos which were recorded in each of the specific topics.  This measure is commonly used for emotion assessment scenarios \cite{Bunney1963,Birmaher1997,BriggsGowan2004} and presents an unbiased measure of agreement. Tables \ref{tab:ICCHHI} and \ref{tab:ICCHRI} exhibit the coefficients for each topic for both scenarios. It is important to notice that the lottery scenario produced better agreement in most cases, and in the food scenario the agreement was worse. Also, the dominance variable was the one with the lowest agreement coefficients, while the emotional concepts had the highest. 
	%
	%
	%
	%
	
	\begin{table}
		\caption[Intraclass coefficient per topic in the HHI scenario.]{Concordance Correlation Coefficient (CCC) per topic in the HHI scenario.}
		\center \begin{tabular}{ c c c c c c}
			\hline
			Characteristic  &         Lottery & Food & School & Family & Pet \\ \hline
			\multicolumn{6}{c}{HHI}\\ \hline
			Valence & 0.7         & 0.5    & 0.3 & 0.6 & 0.4 \\ 
			Arousal  & 0.5   & 0.6 & 0.6 & 0.5 & 0.4 \\  
			Dominance  & 0.4  & 0.5 & 0.4 & 0.5 & 0.4 \\
			Emotion Concept  & 0.7 & 0.6 & 0.5 & 0.6 & 0.5\\\hline
			\multicolumn{6}{c}{HRI} \\ \hline
			Valence & 0.7         & 0.6    & 0.4 & 0.5 & 0.6 \\ 
			Arousal  & 0.6   & 0.6 & 0.6 & 0.4 & 0.5 \\  
			Dominance  & 0.5  & 0.5 & 0.5 & 0.4 & 0.5 \\
			Emotion Concept  & 0.6 & 0.7 & 0.5 & 0.5 & 0.5\\
		\end{tabular} 
		\label{tab:ICCHHI}
	\end{table}
	
	\begin{table}
		\caption[Intraclass coefficient per topic in the HRI scenario.]{Concordance Correlation Coefficient (CCC) per topic in the HRI scenario.}
		\center \begin{tabular}{ c c c c c c}
			\hline
			Characteristic  &         Lottery & Food & School & Family & Pet \\ \hline
			Valence & 0.7         & 0.6    & 0.4 & 0.5 & 0.6 \\ 
			Arousal  & 0.6   & 0.6 & 0.6 & 0.4 & 0.5 \\  
			Dominance  & 0.5  & 0.5 & 0.5 & 0.4 & 0.5 \\
			Emotion Concept  & 0.6 & 0.7 & 0.5 & 0.5 & 0.5\\
		\end{tabular} 
		\label{tab:ICCHRI}
	\end{table}

	\subsection{Unimodal Perception Experiments}
	
	Our model relies heavily on the unisensory pathways to learn expression representations. Thus, we must guarantee that the unisensory channels provide a robust feature-level representation. To evaluate these channels, we use two experiments: one to evaluate the visual channel and one to evaluate the auditory channel.
	
	To evaluate the visual channel we trained and tested our model using the FER+ corpus. This corpus provides its own distribution of training, validation and testing samples, which facilitates the comparison of the results within different classification models. For this experiment, we attach a fully connected hidden layer with 200 units to our  Visual Channel model and use a softmax classifier to categorize the facial expressions. We augment the FER+ training set by using random image translation, rotation, and cropping, so we have more data to train the model. We repeated the experiment 30 times and calculated the mean accuracy for the validation and test set. We fine-tuned the training hyper-parameters of this experiment using the hyperopt library \cite{Bergstra2015}. 
	
	To evaluate the auditory channel we use the Emo-DB corpus. Following the evaluation procedure proposed by the authors of this corpus, we separate randomly 70\% of the data to be used for training, 15\% as validation set used to control the training and 15\% to test the model. We adopt a similar architecture as in our visual channel and connected the auditory channel to a fully connected hidden layer with 200 units and to a softmax classifier. We then repeat the experiments 30 times and calculate the accuracy of the test and validation sets. We also fine-tuned the training hyper-parameter using the hyperopt library.

	\subsection{Crossmodal Perception Experiments}
	
	To evaluate the crossmodal integration and feedback of the Cross-channel we performed experiments using the FER+, Emo-DB and OMG-Emotion corpora. We first pre-trained the cross-channel architecture using both the Emo-DB and the FER+ corpus as described in the unimodal experiments. We then, fine-tuned the architecture using the training set of the OMG-Emotion corpus to train the cross-channel and the last layers of each unimodal pathway. This way, we maintain the robust features which were learned individually but fine-tuned the combined expression descriptor with the cross-channel modulation. Different from the unisensory pathways, which were trained as categorical classifiers, in this experiment we make use of an arousal/valence estimator. For that, we attach a fully connected hidden layer, with 200 units, to two regression layers, one to be trained for arousal and one to be trained for valence. These layers are trained using Mean Square Error (MSE) and output a value between -1 and 1. The idea of using two regression layers instead of one go in the direction of maintaining arousal and valence as independent variables, although being able to be described using the same features. We then evaluate our model using the test and validations set of the OMG-Emotion corpus and calculate the Concordance Correlation Coefficient (CCC) between the annotators and the model's output for both arousal and valence.
	
	We then evaluated the role of the Perception GWR on clustering different expressions. We first pre-train the Perception GWR using the OMG-Emotion training subset. By doing this, the Perception GWR will learn coincident crossmodal stimuli, meaning that a smiling face will be associated with a happy voice, for example. This is an important step, as it will give the Perception-GWR the capability to associate congruent crossmodal stimuli. We then use the same classifier used in the previous experiment to classify the BMUs for each sample on the OMG-Emotion test subset.

	To understand the changes that the cross-channel modulations perform on the expression representation, we experiment with different values for the modulator factor for each channel. We measure this impact using the changes on the CCC and optimize the model to maximize the CCC. The final modulator factor we used was 0.7 for the Visual Channel and 0.4 for the Auditory Channel.
	

	\subsection{Intrinsic Emotion Behavior Experiments}
	
	Once we have a general perception module with the trained CCCNN and Perception GWR, we proceed to evaluate the complete emotion appraisal capabilities of our model to describe the emotion behavior of a participant.  For that, we use the different scenarios and topics of the KT-Emotion Interaction Corpus to generate an intrinsic emotion description, which will be compared with the annotations of the corpus.
	
	We first evaluate how well the framework describes individual user behavior. We use our framework to analyze all the videos of the corpus and create individual Affective Memories for each participant. We then calculate the change of the Affective Memories over time and compare these changes with the annotations for each video using the CCC. With that, we can quantify how similar is the formation and evaluation of the Affective Memory.
	
	Then we run an experiment with the data recorded from the iCub robot. We evaluated the mood formation of the model when interacting with different persons over different topics. This experiment intends to exemplify the use of the mood as an intrinsic and adaptive emotion descriptor, which does not only imitate what was perceived, as the Perception GWR, neither only remember the emotion behavior, as the Affective Memory, but creates a general representation of the human behavior over time take into consideration these two concepts. To demonstrate this effect, we train the model using different persons which performed the interaction on the same topic but we created an individual mood for Human-Human Interaction (HHI) and one for Human-Robot Interaction (HRI). We show the mood formation in each of this scenarios, and we quantify the difference of each scenario by calculating the CCC between each of these moods and the annotated user behavior.

	\section{Results}
	
	\subsection{Unimodal Perception}
	
	The results of training and evaluating our Visual Channel with the FER+ corpus are exhibited in Table \ref{tab:resultsVisionChannel}. It is possible to see that the performance obtained by the Visual Channel is actually very competitive with state-of-the-art models evaluated on the same dataset, which shows an indication that our Visual Channel learns robust facial features for categorical emotion recognition.

	\begin{table}
		\caption{Mean accuracy (\%) and standard deviation the Visual Channel and other models evaluated on the test set of the FER+ corpus.}
		\center \begin{tabular}{ c c}
			\hline
			Model  &         Accuracy (Std) \\ \hline
			Vision Channel & 85.8 \% (1.2) \\
			\cite{Barsoum2016} & 84.9 \% (0.3)\\
			\cite{Chen2017} & 82.8 \% \\
			\cite{Huang2017} & 86.58 \% \\				
			
		\end{tabular} 
		\label{tab:resultsVisionChannel}
	\end{table}
	
	Evaluating our Auditory Channel with the Emo-DB corpus are exhibited on Table \ref{tab:resultsAuditoryChannel}. In this case, it is possible to see that the Auditory Channel has a similar performance as newer models based on deep learning techniques. This also shows that our Auditory Channel is quite competitive with recent state-of-the-art models, and confirms that the learned representations are robust on a unisensory level.

	\begin{table}
		\caption{Mean accuracy (\%) and standard deviation the Auditory Channel and other models evaluated on the test set of the Emo-DB corpus.}
		\center \begin{tabular}{ c c}
			\hline
			Model  &         Accuracy (Std) \\ \hline
			Auditory Channel & 88.7 \% (2.3)  \\
			\cite{Deb2017} & 85.10 \% (-)\\
			\cite{Deb2018} & 83.80\% (-)\\			
			
		\end{tabular} 
		\label{tab:resultsAuditoryChannel}
	\end{table}

	\subsection{Crossmodal Perception}

	The results of our experiments to evaluate the crossmodal perception are reported in Table \ref{tab:resultsCrossmodalPerception}. We first evaluate the capability of the proposed cross-channel architecture to learn robust features. It is possible to see that the pure cross-channel classifier, although produces better results than the baseline, suffers when compared to models which use recurrent connections \cite{peng2018deep,zheng2018multimodal}. This possibly happens due to the OMG-Emotion samples rely heavily on continuous expression representations. 
	
	\begin{table}
		\caption{Mean accuracy (\%) and standard deviation the Cross-channel and other models evaluated on the test set of the OMG-Emotion Recognition corpus.}
		\center \begin{tabular}{ c c c }
			\hline
			Model  &   CCC Arousal & CCC Valence \\ \hline
			PerceptionGWR & 0.38 & 0.47  \\
			Cross-channel & 0.30 & 0.38  \\
			\cite{zheng2018multimodal} & 0.36 & 0.49 \\
			\cite{peng2018deep} & 0.23 & 0.44 \\
			
		\end{tabular} 
		\label{tab:resultsCrossmodalPerception}
	\end{table}
	
	This effect is also the reason why our PerceptionGWR experiment provided better results, reaching state-of-the-art performance on this corpus. By using a recurrent GWR layer, our model was able to learn the temporal characteristics, along with the contextual features of the perceived expressions. It is also important to note that our model, once pre-trained, can learn to cluster the expressions in an unsupervised manner, which means that the same model can now be adapted easily to different domains.

	\subsection{Intrinsic Emotion Behavior}
	
	Training the affective memory of our Emotional Deep Neural Circuitry gave us a direct correlation between what was expressed and the stored memory. This means that the memory learned how to create neurons that will code for the presented emotional concepts. The correlation coefficients calculated for the HHI scenario are shown in Table \ref{tab:ICCHHISubjectNetwork}. Here, it is possible to see that for most of the subjects, the network presented a slightly good correlation, while only a few presented a very good one. 
	
	\begin{table}
		\caption[Concordance Correlation Coefficient (CCC) of our model per subject on the HHI scenario.]{Concordance Correlation Coefficient (CCC) of our Affective Memories per subject in the HHI scenario.}
		\center \begin{tabular}{ c c c c c c c c c}
			\hline
			Session &  \multicolumn{2}{c}{2} & \multicolumn{2}{c}{3} & \multicolumn{2}{c}{4} &  \multicolumn{2}{c}{5} \\ \hline
			Subject  &         S0 & S1 & S0 & S1 & S0 & S1 & S0 & S1 \\ \hline
			Valence & 0.63         & 0.54    & 0.67 & 0.59 & 0.69 & 0.67 & 0.54 & 0.59 \\ 
			Arousal  & 0.55   & 0.57 & 0.67 & 0.59 & 0.67 & 0.60 & 0.57 & 0.53 \\  
			Emotional Concepts  & 0.79 & 0.67 & 0.74 & 0.79 & 0.61 & 0.74 & 0.67 & 0.59\\  \hline
			
			Session &  \multicolumn{2}{c}{6} & \multicolumn{2}{c}{7} & \multicolumn{2}{c}{8} \\ \hline
			Subject  &         S0 & S1 & S0 & S1 & S0 & S1  \\ \hline
			Valence & 0.57         & 0.61    & 0.64 & 0.61 & 0.49 & 0.68  \\ 
			Arousal  & 0.54   & 0.61 & 0.50 & 0.87 & 0.71 & 0.84 \\ 
			Emotional Concepts  & 0.68 & 0.87 & 0.68 & 0.63 & 0.64 & 0.76 \\  \hline 
			
		\end{tabular} 
		\label{tab:ICCHHISubjectNetwork}
	\end{table}
	
	For the subjects in the HRI scenario, the correlation coefficients are presented in Table \ref{tab:ICCHRISubjectNetwork}. It is possible to see that there is a drop in the concordance when compared to the HHI scenario. This happens possibly by the differences while performing the dialogue between interacting with a human and interacting with a robot. Usually, when interacting with robots, humans tend to be less expressive and more dubious.
	
	\begin{table}
		\caption[Intraclass coefficient of our model per subject on the HRI scenario.]{Concordance Correlation Coefficient (CCC) of our Affective Memories per subject in the HRI scenario.}
		\center \begin{tabular}{ c c c c c c c c c c}
			\hline
			Subject  &         S1 & S2 & S3 & S4 & S5 & S6 & S7 & S8 & S9 \\ \hline
			Valence & 0.58         & 0.42    & 0.67 & 0.59 & 0.42 & 0.80 & 0.45 & 0.61& 0.78\\ 
			Arousal  & 0.52   & 0.62 & 0.57 & 0.60 & 0.67 & 0.62 & 0.59 & 0.48& 0.58 \\  
			Emotional Concept  & 0.74 & 0.57 & 0.62 & 0.61 & 0.57 & 0.59 & 0.57 & 0.69& 0.72\\

		\end{tabular} 
		\label{tab:ICCHRISubjectNetwork}
	\end{table}
	
	The results of evaluating our model for the intrinsic mood formation for the HHI scenario are presented in Table \ref{tab:ICCHHINetwork}. It is possible to see a very high correlation for both dimensions, valence, and arousal, for at least two scenarios: Lottery and Food. These two scenarios were the ones with a stronger correlation also within the annotators, and possibly the ones where the expressions were most easily distinguishable for all the subjects. 
	
	\begin{table}
		\caption[Intraclass coefficient of our model per topic on the HHI scenario.]{Concordance Correlation Coefficient (CCC) of our intrinsic mood per topic in the HHI scenario.}
		\center \begin{tabular}{ c c c c c c}
			\hline
			Characteristic  &         Lottery & Food & School & Family & Pet \\ \hline
			Valence & 0.65         & 0.64    & 0.41 & 0.67 & 0.57 \\ 
			Arousal  & 0.67   & 0.72 & 0.42 & 0.56 & 0.49 \\  
			Emotional Concept  & 0.84 & 0.71 & 0.47 & 0.52 & 0.53\\  
		\end{tabular} 
		\label{tab:ICCHHINetwork}
	\end{table}
	
	The correlation coefficients for the HRI scenario are presented in Table \ref{tab:ICCHRINetwork}. It is possible to see that, similarly to the HHI scenario, the topics with the highest correlation were Lottery and Food, while the lowest ones were Pet and Family. Here the correlation values are slightly smaller than in the HHI scenario, indicating that for these expressions were more difficult to describe, which is a behavior inherited from the Affective Memory module.
	
	\begin{table}
		\caption[Intraclass coefficient per topic in the HRI scenario.]{Concordance Correlation Coefficient (CCC) of our intrinsic mood per topic in the HRI scenario.}
		\center \begin{tabular}{ c c c c c c}
			\hline
			Characteristic  &         Lottery & Food & School & Family & Pet \\ \hline
			Valence & 0.78         & 0.67    & 0.31 & 0.58 & 0.47 \\ 
			Arousal  & 0.72   & 0.61 & 0.57 & 0.49 & 0.42 \\  
			Emotion Concept  & 0.79 & 0.75 & 0.62 & 0.51 & 0.57\\
		\end{tabular} 
		\label{tab:ICCHRINetwork}
	\end{table}
	
	\section{Discussions}
	
	The obtained results demonstrate that our model is able to both describe individual emotion perception and to describe emotion behavior over time. This is possible due to our adaptive learning, which is able also to adapt to different persons and emotion expressions continuously. By updating the affective memories over a long time span, the robot can have an insight into the behavior of that specific person. This is much necessary for personalized interaction with different users. 
	
	To be able to describe long-term emotional behavior from personalized interactions, our model needs a very strong basis for emotion expression recognition. Our results demonstrate that the model for external emotion appraisal is very robust for spontaneous, unimodal and multimodal emotion expressions, and is competitive with different state-of-the-art models. This is of extreme importance for the internal emotion appraisal model, composed of the mood GWR and affective memories, as they rely drastically on the external appraisal model. Our internal appraisal model presents a completely different behavior from commonly used emotion recognition models, being able to describe personalized long-term emotion behavior, and we recognize it as the main contribution of this work.
	
	Two aspects of our model, however, are demonstrated using objective measures. First, is the capability of the model to describe the emotion behavior of a person over time. This is an important aspect of our model that allows the robot to maintain a state of the person's behavior and at the same time identify which direction the interaction is moving to. The second aspect is the impact of the modulation while forming the intrinsic mood. This session will discuss in details these two aspects using demonstrative examples of each of them.

	\subsection{Understanding Affective Behavior Through Time}
	
	By using a growing network to represent our Affective Memories, we are able to store not only how a certain participant behave in the past, but also to have a certain direction on how this behavior changed. By observing the age of each neuron in our Affective Memory, we are able to identify when a certain affective behavior happened and describe, time-wise, the past and current interactions. 
	
	\begin{figure} 
		\center{\includegraphics[width=1\linewidth]{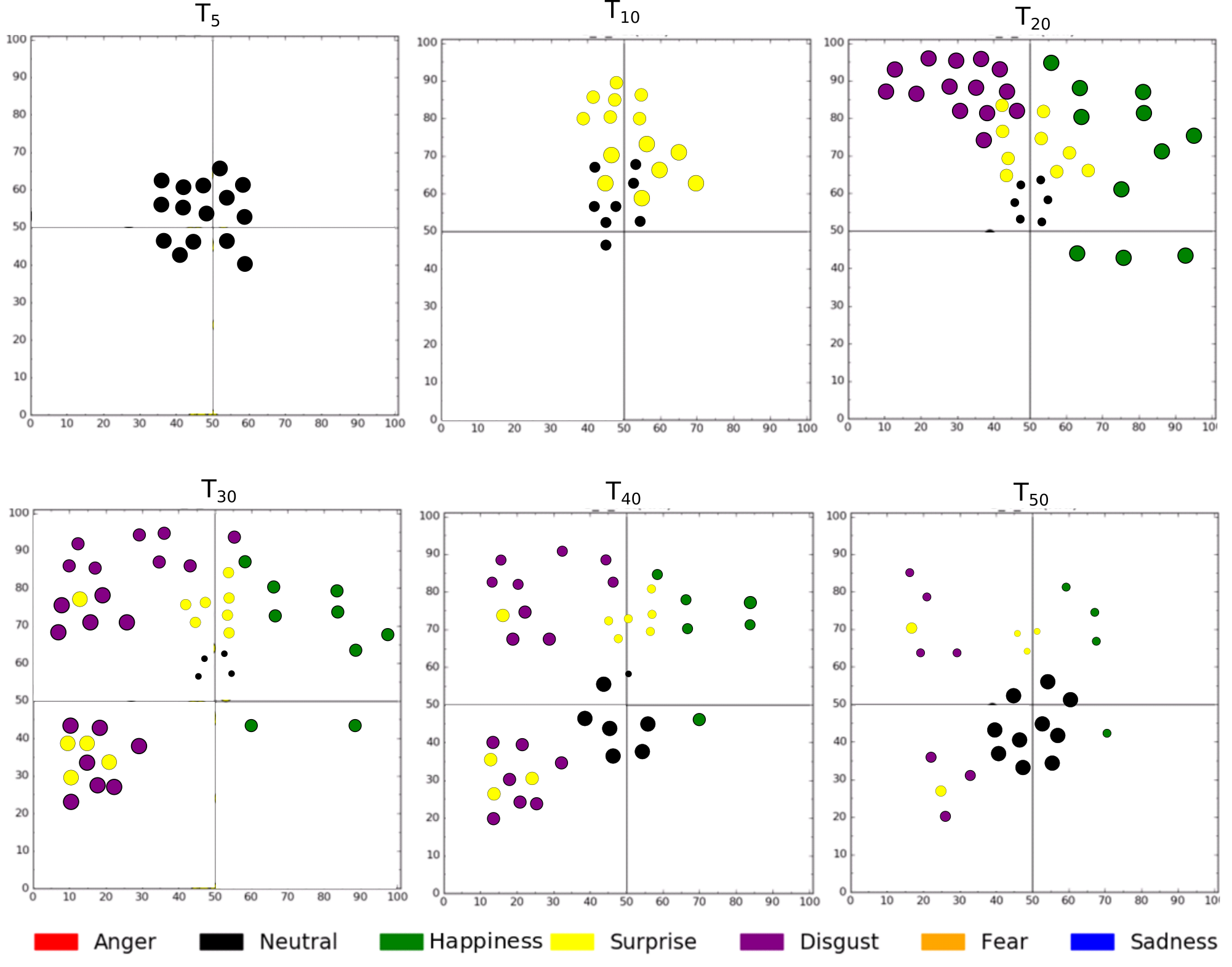}}
		\caption[Illustration of the effects of memory modulation on HHI scenario.]{Illustration of the neural growth of the mean of different Affective Memories for the 'Food" topic. Each plot demonstrates the arousal/valence axis of the neurons for each interaction time(starting from 5 seconds, and going until 50 seconds). In each plot, the size of the neuron represents the neuron's age.}
		\label{fig:timePerception}
	\end{figure}
	
	In the example illustrated in Figure \ref{fig:timePerception}, we can see the mean behavior of different Affective Memories for each participant on the ''Food" topic. It is possible to see that at the beginning of the interaction, before the first 5 seconds, most of the participants demonstrated a neutral behavior. As soon as the robot starts speaking, the participants demonstrated a high-arousal surprise behavior. This probably happens as most of the participants never interacted with the iCub before. As soon as the robot starts to present the topic, proposing the participants to eat a disgusting food (around 20s), some participants demonstrate a clear high-arousal disgust expression, while other stated laughing and were captured as high-arousal happy expressions. While the interaction goes (in between the first 30s and 40s of interaction), neurons representing surprise expressions start to be forgotten, while low-arousal neurons representing disgust appears. That probably is explained by the participants processing the information and getting into the scenario. Finally, around interaction 40s and 50s of interaction, the topic did not affect the participants anymore, and most of them start to display neutral expressions. That caused new neutral neurons to appear, while the happy and mostly surprised neurons slowly being forgotten.
	
	This behavior of our model is a very important characteristic that will allow the robot to take actions considering the previous state of the interaction. Using our adaptive system we are able to describe the interaction with details in a neural representation which is robust and concise. 
	
	\subsection{The Role of Modulation}
	
	In our experiments, we evaluated how the network behaves in different scenarios: from unisensory perception to formulation of an intrinsic emotion appraisal. One interesting aspect of our model is the ability to use the memory modulation during the intrinsic mood generation. Although very difficult to evaluate objectively, the modulation has an important role in enforcing the mood generation based on what was perceived. Without the use of the modulation, the mood would be equally influenced by all the neurons on the  memory, which would mean that the current contextual information has no effect on it.
	
\begin{figure} 
		\center{\includegraphics[width=1\linewidth]{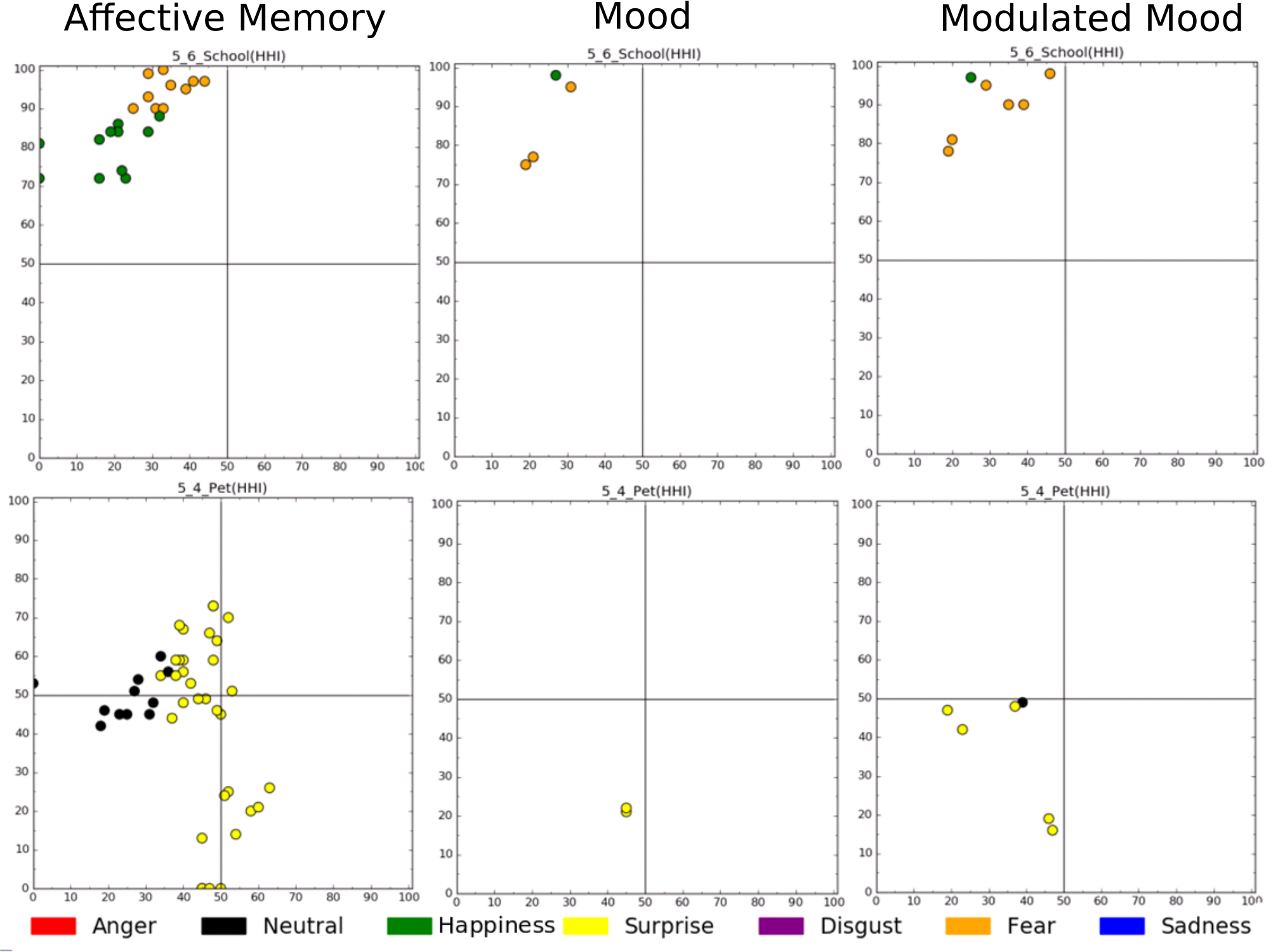}}
		\caption[Illustration of the effects of memory modulation on HHI scenario.]{Illustration of the effects of memory modulation from the Mood Memory in the Affective Memory for two subjects of the HHI experiment in the Food and Pet scenarios.}
		\label{fig:memoryModulationHHI}
	\end{figure}

	To evaluate this effect, we present to the network one full dialogue interaction to the model, containing the passive subject and one topic. We proceed with the evaluation on two subjects, 6 and 4, of the HHI scenario, with videos for the topics "School" and "Pet". Figure \ref{fig:memoryModulationHHI} illustrates the arousal, valence and emotional concepts of the neurons for the Affective Memory, Mood Memory and an Affective Memory with mood modulation for this experiment.

	It is possible to see that the Mood Memory contains much less information, however, code for the general behavior of the scene. On subject 6 it is possible to see that the Mood Memory codes information with very similar arousal, valence, and emotional concepts. When used as a modulator, what happens is that the amount of information decreases drastically, however, the structure and amplitude of the three dimensions do not change much.
	
	We repeat the same investigation with subjects 3 and 7 for the Food and Pet dialogues of the HRI scenario. It is possible to see again how the mood was affected by the perceived information, while not changing much its own topological structure. In the Pet topic, the Affective Memory module contains very little coded information, and thus had a stronger effect on the modulated mood. This happens because this interaction was much shorter than the others, presenting to the network fewer expressions. When a larger amount of expressions are present, as in the Food topic, the network tends to behave differently, as the mood has more information to update.
	
	\begin{figure} 
		\center{\includegraphics[width=1\linewidth]{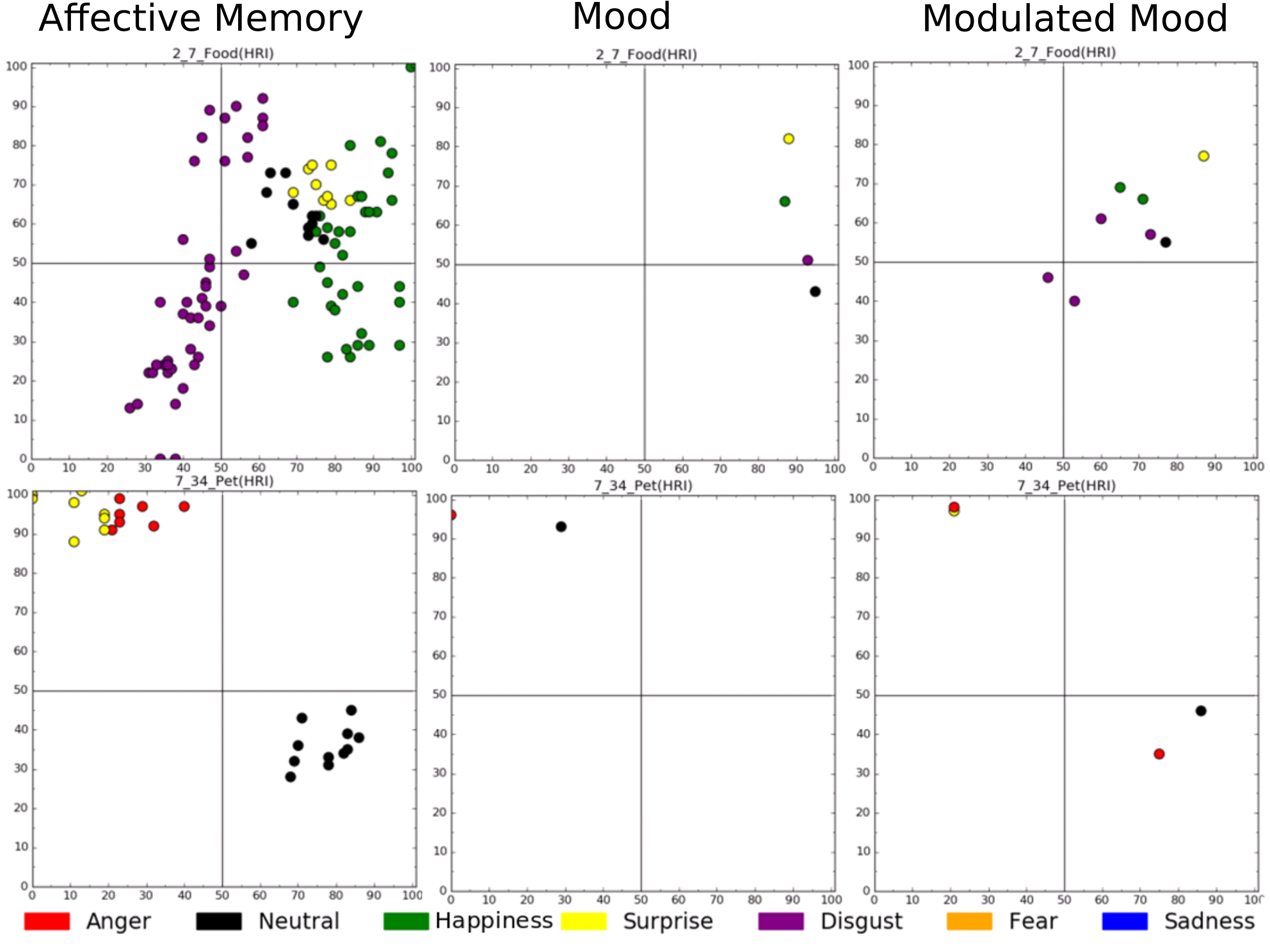}}
		\caption[Illustration of the effects of memory modulation on HRI scenario.]{Illustration of the effects of memory modulation from the Mood Memory in the Affective Memory for two subjects of the HRI experiment in the Food and Pet scenarios.}
		\label{fig:memoryModulationHRI}
	\end{figure}

	\section{Conclusion}
	
	Emotions are present in many aspects of our lives, that include interpersonal communication, learning and experiencing new things, and  remembering past events. Our understanding of emotions is yet to develop, however we are certain on the necessity to approach this topic, ranging from philosophy to neurosciences, in a multidisciplinary way. In particular, embedding concepts and ideas related to emotions in autonomous intelligent systems is still an open field, with several different problems to be solved. The proposed model addressed some of these problems, by using state of the art hybrid neural network inspired by behavioral and neural mechanisms in humans. 
	
	We proposed an emotion appraisal model, which can categorize in context the perceived expressions by a robot and create an intrinsic mood in line with the framework of developmental robotics. The model consists of a perception layer, implemented as a Cross-channel Convolution Neural Network. We propose conceptual updates on this model and introduce the use of modulating connections, which improve the model's stability to deal with unisensory and multisensory stimuli. The proposed PerceptionGWR model implements a recurrent growing-when-required network which can learn to cluster emotion concepts when perceived, and thus, presents a continuous learning model for emotion expressions. This is a step forward to the model of Barakova et al. \cite{Barakova2015} which although models reactions to emotional events does not allow for learning and thus has limits for implementation on artificial systems such as social robots. Finally, our intrinsic mood representation and affective memory module work together to create an internal appraisal module. This module presents a continuous formation and update of the emotional concepts, which can be used to describe different emotional states based on what was perceived.
	
	Our experiments show that the proposed model is competitive with state-of-the-art models for emotion perception, and it is able to describe long-term emotional behavior, in the scale of several minutes. To evaluate it, we proposed a novel corpus composed of human-human and human-robot interactions. This corpus is the first to include, in a controlled scenario but with loose interaction constrictions, material for direct comparison on how different humans interact with other humans or with robots. We use the proposed model to create intrinsic descriptions of what was perceived, simulating the mood of a robot.
		
	The analysis of our model representations could be expanded in several different ways. We encourage both research and applications on how our model would behave to different scenarios. The development of such model is not bounded by any constraint, and the updates of different concepts proposed here, if beneficial to the model, are also encouraged.
		
		\section*{Acknowledgements}
		This work was partially supported by the German Research Foundation (DFG) under project CML (TRR 169) and the NSFC (61621136008) and the China Scholarship Council. The authors would like to thank German I. Parisi and Katja Koesters for important suggestions and support on the development of our model and manuscript.

		\section*{Bibliography}

		\bibliographystyle{spmpsci}  
		\bibliography{bib}
		
	\end{document}